\documentclass[10pt,twocolumn,letterpaper]{article}

\usepackage[accsupp]{axessibility} % Improves PDF readability for those with visual impairments.
\usepackage{cvpr}              % To produce the CAMERA-READY version
\usepackage{multirow}
\usepackage{colortbl}

\usepackage[dvipsnames]{xcolor}
\definecolor{cvprblue}{rgb}{0.21,0.49,0.74}
\definecolor{pink}{rgb}{1.0,0,1.0}
\usepackage[pagebackref,breaklinks,colorlinks,citecolor=cvprblue]{hyperref}
\usepackage{tikz}
\usepackage{graphicx}
\usepackage{lipsum}
\newcommand\blfootnote[1]{%
  \begingroup
  \renewcommand\thefootnote{}\footnote{#1}%
  \addtocounter{footnote}{-1}%
  \endgroup
}

\usepackage[safe]{tipa}
\usepackage{xspace}
\usepackage[dvipsnames]{xcolor}
\def\paperID{8464} % *** Enter the Paper ID here
\def\confName{CVPR}
\def\confYear{2024}

\newcommand{\myparagraph}[1]{\noindent\textbf{#1}\,\,}

\title{Stratified Avatar Generation from Sparse Observations}
 \author{
Han Feng$^{1*,\ddag}$ \quad Wenchao Ma$^{2*}$ \quad Quankai Gao$^{3}$ \quad Xianwei Zheng$^{1}$ \quad Nan Xue$^{4\dag}$ \quad Huijuan Xu$^{2}$ \\[5pt]
$^1$Wuhan University \qquad 
$^2$Pennsylvania State University\\% \quad 
$^3$University of Southern California \qquad
$^4$Ant Group\\
 Project website: \href{https://fhan235.github.io/SAGENet/}{\textcolor{pink}{https://fhan235.github.io/SAGENet/}}
}

\begin{document}
%\maketitle

%\begin{figure*}[ht!]
%    \centering
%    \includegraphics[width=0.99\linewidth]{img/teaser.pdf}
%    \caption{Teaser}
%    \label{fig:enter-label}
%\end{figure*}

\twocolumn[{%
\renewcommand\twocolumn[1][]{#1}%
\maketitle
\thispagestyle{empty}

\begin{center}
    % \centering
    % \captionsetup{type=figure}
    % \subfloat[Sparse Observation\label{fig:sp-input}]{\includegraphics[width=0.24\linewidth]{img/teaser/sparse.pdf}}
    % \subfloat[Upper Body Reconstruction\label{fig:upper-recon}]{\includegraphics[width=0.24\linewidth]{img/teaser/upper.pdf}}
    % \subfloat[Lower Body Reconstruction\label{fig:lower-recon}]{\includegraphics[width=0.24\linewidth]{img/teaser/lower.pdf}}
    % \subfloat[Full Body Reconstruction\label{fig:full-recon}]{\includegraphics[width=0.24\linewidth]{img/teaser/full.pdf}}
    % \includegraphics[width=\linewidth]{img/teaser.pdf}
    \begin{minipage}[t]{0.19\linewidth}
            \centering
            \includegraphics[width=0.99\linewidth]{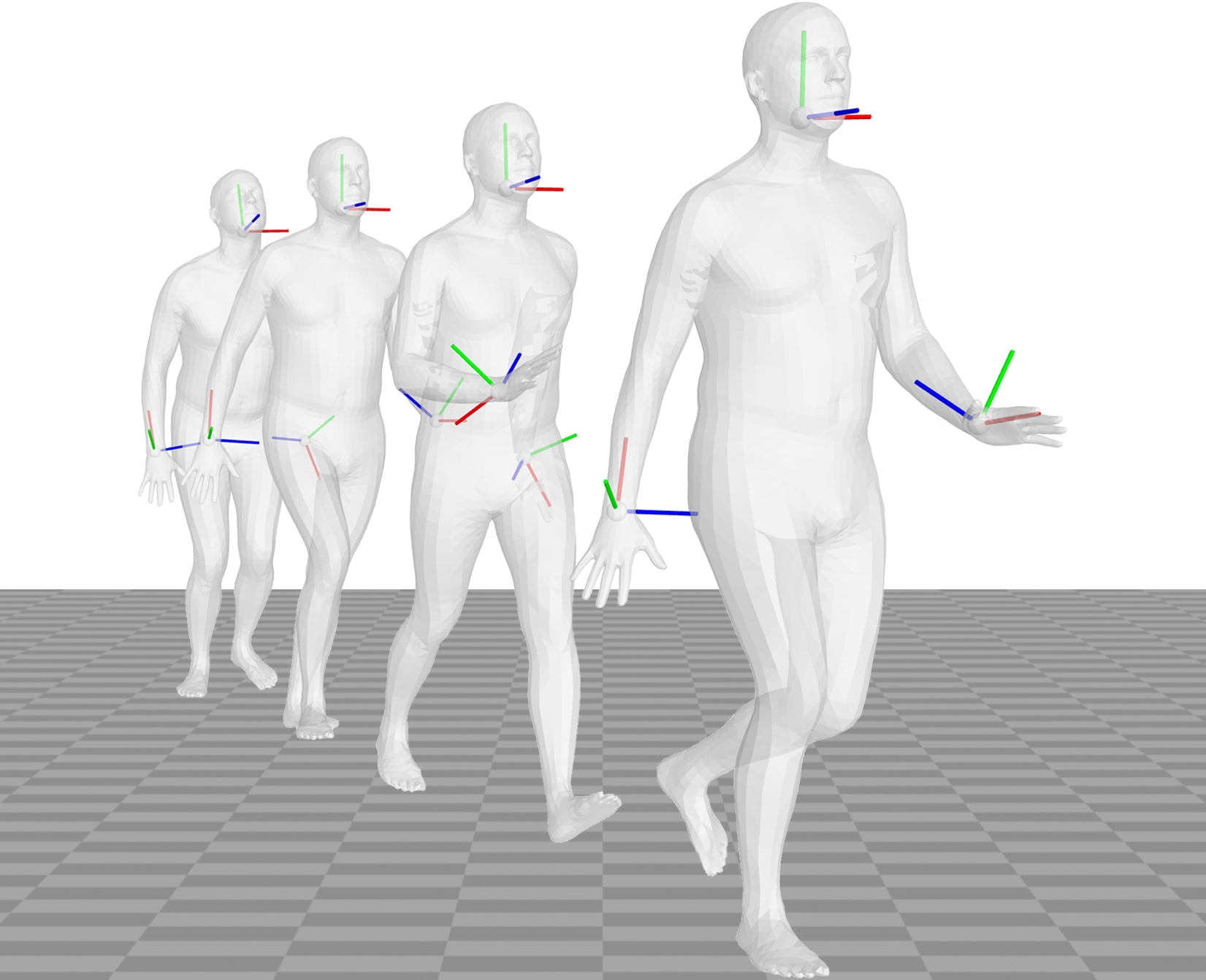}\\
            \small{(a) Sparse Observation}
    \end{minipage}
    \begin{minipage}[t]{0.19\linewidth}
        \centering
        \includegraphics[width=0.99\linewidth]{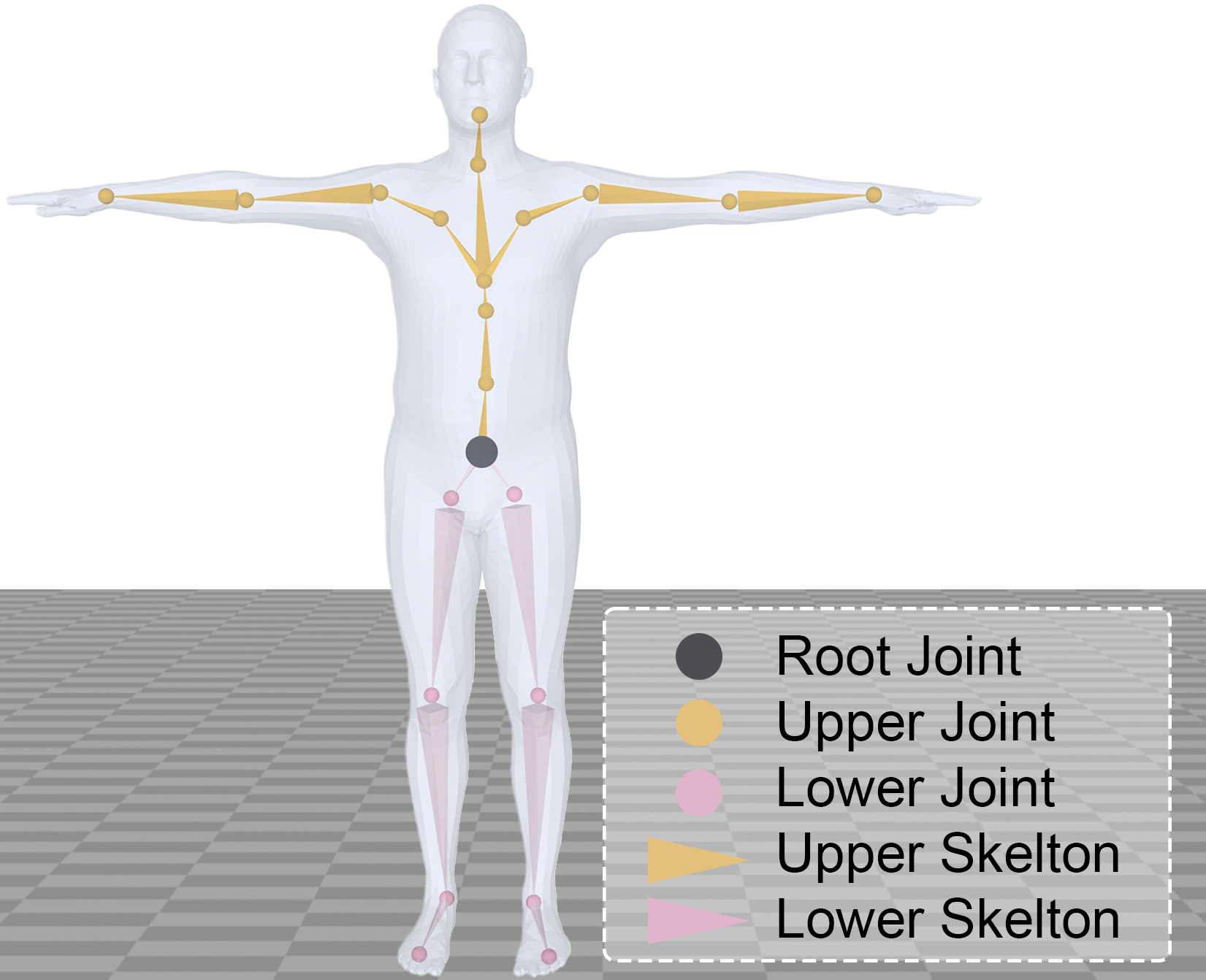}
        \\ \small{(b) Disentangled Body}
    \end{minipage}
    \begin{minipage}[t]{0.19\linewidth}
        \centering
        \includegraphics[width=0.99\linewidth]{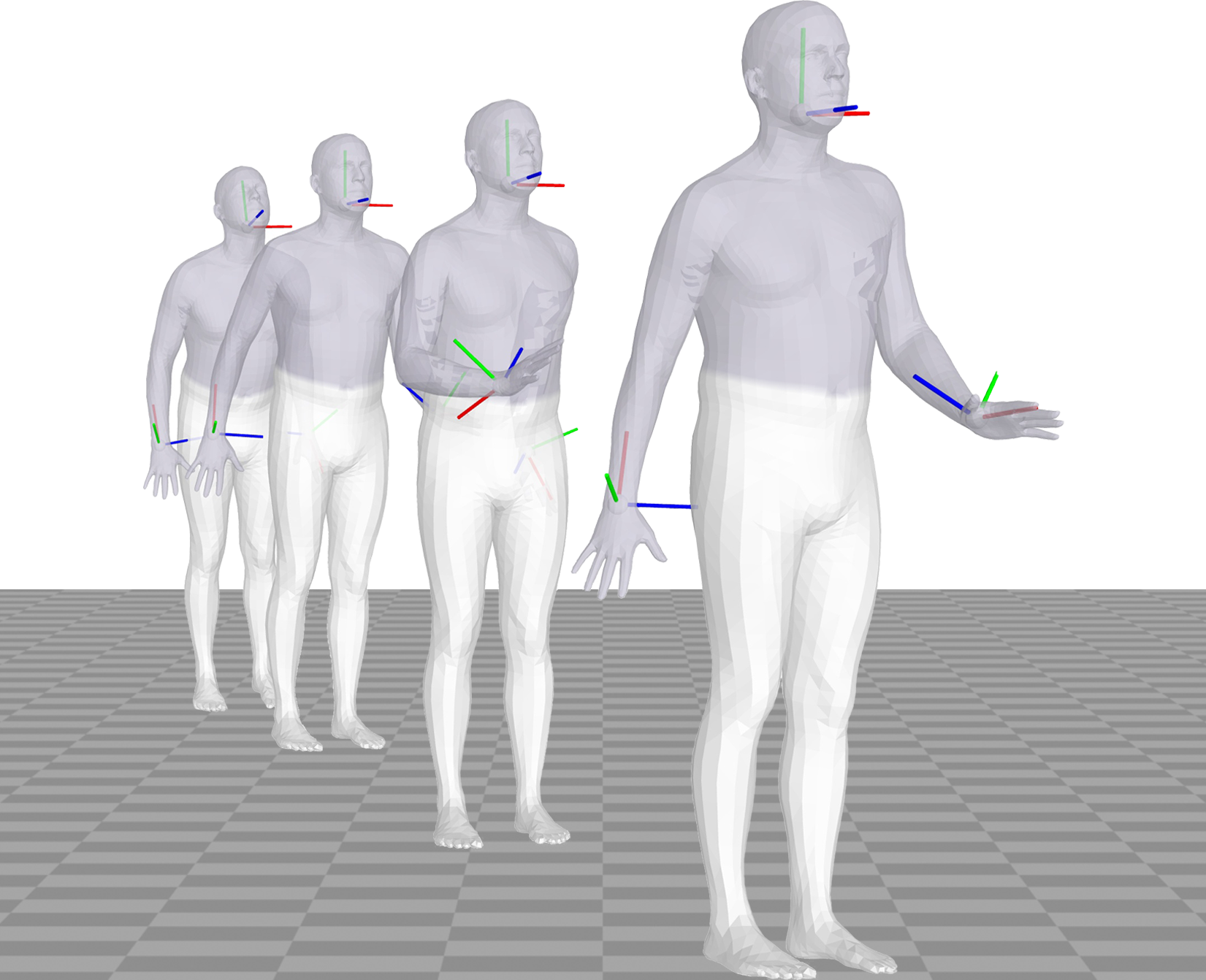}\\
        \small{(c) Upper-Body Recon.}
    \end{minipage}
    \begin{minipage}[t]{0.19\linewidth}
        \centering
        \includegraphics[width=0.99\linewidth]{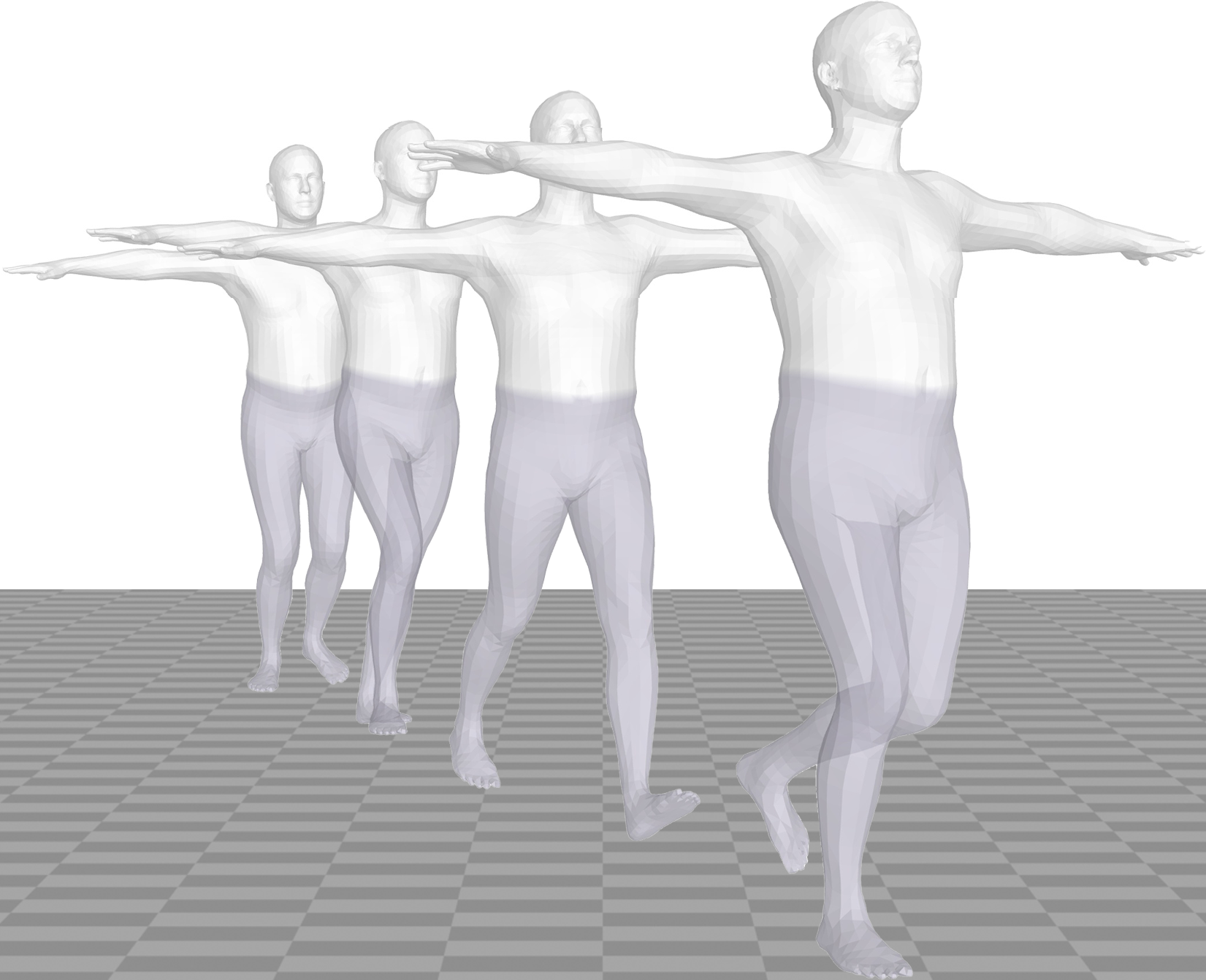}\\
        \small{(d) Lower-Body Recon.}
    \end{minipage}
    \begin{minipage}[t]{0.19\linewidth}
        \centering
        \includegraphics[width=0.99\linewidth]{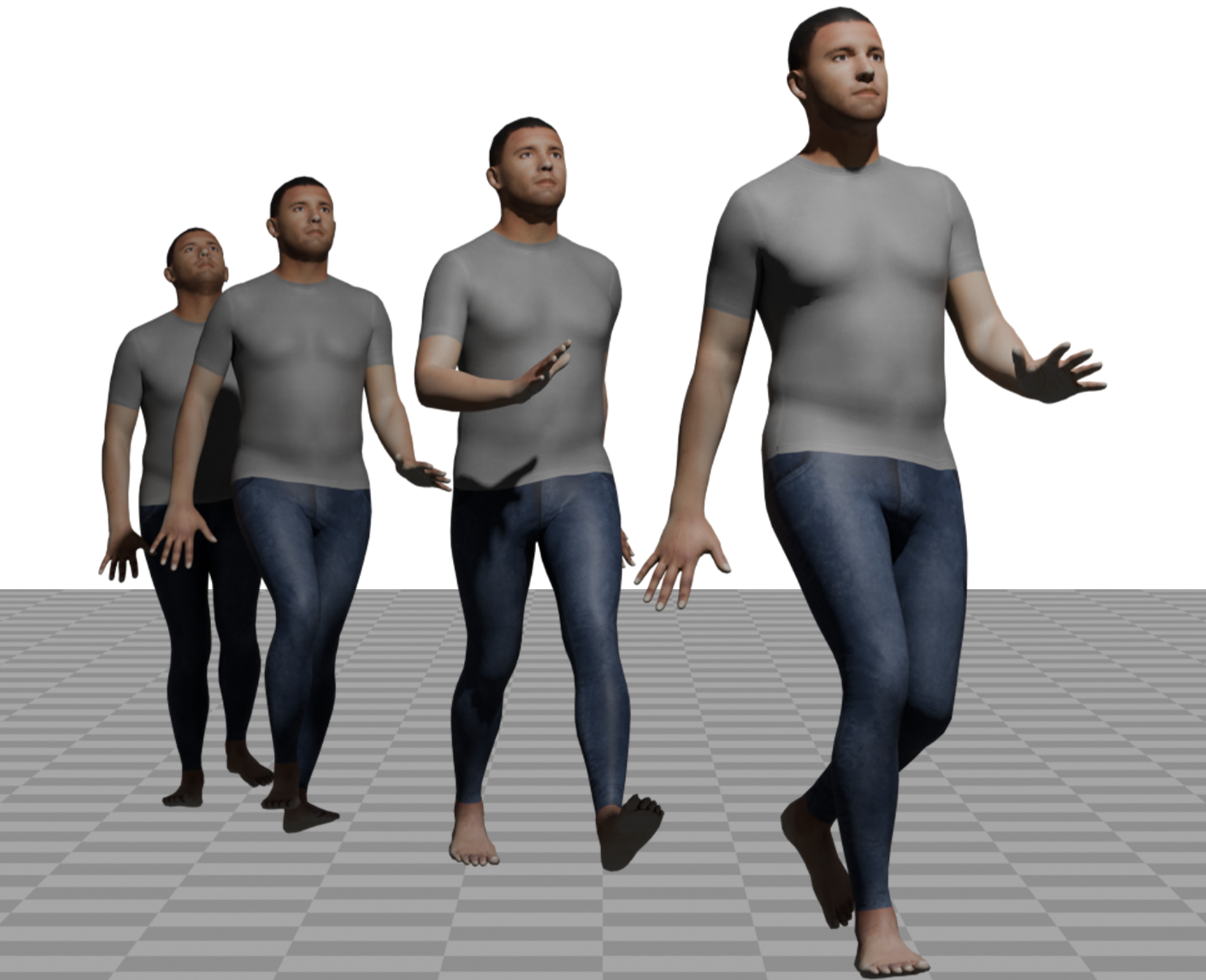}\\
        \small{(e) Full-Body Recon.}
    \end{minipage}
    \captionof{figure}{{\bf Stratified avatar generation from sparse observations}. Given the sensory sparse observation of the body motion: 6-DoF poses of the head and hand marked by RGB axes in (a), our method leverages a disentangled body representation in (b) to reconstruct the upper-body conditioned on the sparse observation in (c), and lower-body conditioned on the upper-body reconstruction in (d) to accomplish the full-body reconstruction in (e).
    % (a): we first disentangle the SMPL body model into upper and lower parts, connected by the root joint. The orientation of the head and hand marked by RGB axes is the sparse inputs to the model shown in (b), and the model output is the full-body reconstruction shown in (e). (c), (d) and (e) show the process of stratified avatar generation, from upper-body reconstruction to lower-body reconstruction and a final full-body reconstruction, with the darker part highlighting the generation outcome from that stage.
    }
    % \vspace{-3mm}
    \label{fig:teaser}
\end{center}%
}]

\blfootnote{\leftline{* equal contributions, ordered by alphabet.}}
\blfootnote{\leftline{$\dag$ Corresponding author: \url{xuenan@ieee.org}}}
\blfootnote{\leftline{$\ddag$ indicates the School of Computer Science, Wuhan University}}

\vspace{-3mm}
\begin{abstract}
\vspace{-3mm}
Estimating 3D full-body avatars from AR/VR devices is essential for creating immersive experiences in AR/VR applications. This task is challenging due to the limited input from Head Mounted Devices, which capture only sparse observations from the head and hands. Predicting the full-body avatars, particularly the lower body, from these sparse observations presents significant difficulties. In this paper, we are inspired by the inherent property of the kinematic tree defined in the Skinned Multi-Person Linear (SMPL) model, where the upper body and lower body share only one common ancestor node, bringing the potential of decoupled reconstruction. We propose a stratified approach to decouple the conventional full-body avatar reconstruction pipeline into two stages, with the reconstruction of the upper body first and a subsequent reconstruction of the lower body conditioned on the previous stage. 
To implement this straightforward idea, we leverage the latent diffusion model as a powerful probabilistic generator, 
and train it to follow the latent distribution of decoupled motions explored by a VQ-VAE encoder-decoder model. Extensive experiments on AMASS mocap dataset demonstrate our state-of-the-art performance in the reconstruction of full-body motions.

\end{abstract}    
\vspace{-2mm}
\section{Introduction}
\label{sec:intro}
Generating 3D full-body avatars from observations of Head Mounted Devices (HMDs) is crucial for enhancing immersive AR/VR experiences. HMDs primarily track the head and hands, while leaving the rest of the body unmonitored. 
This limited motion tracking poses a challenging scenario for accurately reconstructing full-body 3D avatars, particularly in representing the lower body. 
The high degree of freedom in body movements compounds this difficulty, making the task of reasoning human motion from such sparse observations significantly complex.

Tremendous efforts have been made to obtain more tracking signals by adding sensors at Pelvis~\cite{FLAG,VAEHMD,Humor} or both Pelvis and Legs~\cite{SIP,DIP,Transpose,PIP,TIP}.  While these approaches provide more data points for avatar construction, they can diminish the user's experience. Wearing extra devices can be cumbersome, potentially interfering with the user's comfort and immersion in the virtual environment. This trade-off highlights the need for innovative solutions that can deliver detailed body tracking without compromising the user's comfort and immersion in AR/VR settings. Accordingly, we are interested in the problem of generating 3D full-body avatars from sparse observations of HMDs that track the motion of the head and two hands, by developing a neural solution that learns the distribution of full-body poses given the sparse observations as the input condition. 

Recent studies have attempted to address the challenge of sparse observations in HMD-based full-body avatar generation by employing regression-based techniques, as seen in~\cite{AvatarPoser,AvatarJLM}, or by adopting generation-based approaches like~\cite{BoDiffusion,AGROL}. These methods typically use deep neural networks to predict human motion within a single, expansive motion space. However, due to the limited data provided by sparse observations, these networks often struggle to fully capture the complexities of human kinematics across such a broad and unified motion space. This limitation frequently results in reconstructions that are unrealistic and lack physical plausibility.

% \textcolor{red}{Taking the fact of inherent complexities of body motion, we propose a stratified approach to human motion approach, Stratified Avatar Generation (SAGE), which disentangles the upper and lower-body motions and learn the 3D human motion in a stratified manner, as shown in~\cref{fig:teaser}.  The discretization of latent representations brings us additional benefits through quantization constrains the human body model parameters to a finite, meaningful set of points, further sharpening the model's focus on viable poses.
% As for the  [Wenchao: the original paragraph]}

We introduce a new method for reconstructing full-body human motions from sparse observations, called Stratified Avatar Generation (SAGE). Instead of the upper-body motion prediction that has tracking signals of certain upper joints from sparse observations, predicting lower-body motion is not straightforward as no direct tracking signals about any lower-body joint is given. It is noteworthy that SMPL model~\cite{SMPL} connects the upper and lower half-body by a single root joint, as shown in \cref{fig:teaser} (b), which motivates us to split the full-body motions into upper and lower half-body parts. The benefits are two-fold: 1) the smaller search space achieved by disentanglement facilitates learning and prediction; % by allowing deep models to focus on a more limited set of movements and interactions; 
2) our stratified design makes the modeling and inferring for lower-body motions more accurate and visually appealing by explicitly modeling the correlation and constraint between two half-body motions. To this end, we use VQ-VAE~\cite{VQ-VAE} to encode and reconstruct upper and lower body motions separately. 

%This focus helps in sharpening the accuracy of our model in predicting realistic human motions with meaningful latent body representation and turns the problem of full-body avatar generation to be flexible in learning-based and generative paradigms. 

%these networks often face difficulties in fully grasping the complexities of human kinematics across such a broad and unified motion space.

% Taking the fact of inherent complexities of body motion, we propose a stratified approach to human motion approach, Stratified Avatar Generation (SAGE), which disentangles the upper and lower body motions and learn the 3D human motion in a stratified manner, as shown in~\cref{fig:teaser}. Firstly, we revisit the definition of SMPL model~\cite{SMPL} and observed that the root joint is the only common parent linking the upper and lower body segments. This structural insight enables the effective disentanglement of human motion representation into distinct upper and lower body parts. The smaller space achieved by disentanglement facilitates easier learning and prediction by allowing deep models focus on a more limited set of movements and interactions. We employ the VQ-VAE framework~\cite{VQ-VAE}, which incorporates a quantization operation in an auto-encoder structure, to learn discrete latents for the separately analyzed upper and lower body motions. The discretization of latent representations brings us additional benefits through quantization constrains the human body model parameters to a finite, meaningful set of points, further sharpening the model's focus on viable poses.

With the disentangled latent representation of the upper and lower body motions, we aim to recover the accurate full-body motions from sparse observations with a body-customized latent diffusion model (LDM)~\cite{LatentDiffusion} in a stratified manner. Specifically, as shown in \cref{fig:teaser} (c), \cref{fig:teaser}(d), and \cref{fig:teaser}(e), we first find the latent of upper-body motion condition on the sparse observations (i.e., tracking signals of the head and hands in \cref{fig:teaser}(a)). Then, the latent of lower-body motion is inferred condition on both the predicted upper-body latent and sparse observations. Finally, a full-body decoder takes the two half-body latents as input and outputs the full-body motion.

%benefiting from our disentangled design, we are able to use {\em more} signals than prior arts to address the challenging issue of lower-body motion estimation, of which the process is similar to the upper-body part, but with an additional yet informative upper body motion as the input (\cref{fig:teaser} (c)). Finally, a full-body decoder fuse the outputs from previous modules to reconstruct the full-body motion (\cref{fig:teaser} (d)). 

% With the quantized body representation in latents, we focus on the interplay among the sparse input signals, the upper body segment and the lower one, empowered by body-customized latent diffusion model (LDM)~\cite{LatentDiffusion} in a stratified in both training and test. Specifically, we first infer the upper-body latents from the sparse tracking signals of the head and hands as shown in \cref{fig:sp-input}, then leverage LDM to trace the conditional distribution of latent upper-body motion in the diffusion process, and finally decode the denoised latent in the upper body motion in \cref{fig:upper-recon}. Benefitting from powerful LDM and our disentangled design, we are able to use {\em more} signals than prior arts to address the challenging issue of lower-body motion estimation, of which the process is similar to the upper-body part, but with an additional yet informative enough tracking signal as the input. 

In the experiments, we comprehensively justified our intuitive design of disentangling the upper and lower body motion in a stratified manner. On the large-scale motion capture benchmark AMASS~\cite{AMASS}, our proposed SAGE is exhibiting superior performance in different evaluation settings and particularly performs well in terms of the evaluation metrics for lower-body motion estimation compared to previous state-of-the-art methods.

\section{Related Work}
\label{sec:relatework}
\subsection{Motion Reconstruction from Sparse Input}
The task of reconstructing full human body motion from sparse observations has gained significant attention in recent decades within the research community~\cite{SIP, DIP, Transpose, PIP, TIP, FinalIK, CoolMoves, LoBSTr, VAEHMD, FLAG, AvatarPoser, QuestSim, AGROL, BoDiffusion, AvatarJLM}. For instance, recent works~\cite{SIP,DIP,Transpose,PIP,TIP} focus on reconstructing full body motion from six inertial measurement units (IMUs). SIP~\cite{SIP} employs heuristic methods, while DIP~\cite{DIP} pioneers the use of deep neural networks for this task. PIP~\cite{PIP} and TIP~\cite{TIP} further enhance performance by incorporating physics constraints. With the rise of VR/AR applications, researchers turn their attention toward reconstructing full body motion from VR/AR devices, such as head-mounted devices (HMDs), which only provide information about the user's head and hands, posing additional challenges. LoBSTr~\cite{LoBSTr}, AvatarPoser~\cite{AvatarPoser}, and AvatarJLM~\cite{AvatarJLM} approach this task as a regression problem, utilizing GRU~\cite{LoBSTr} and Transformer Network~\cite{AvatarPoser, AvatarJLM} to predict the full body pose from sparse observations of HMDs. Another line of methods employs generative models~\cite{VAEHMD, FLAG, AGROL, BoDiffusion}. For example, VAEHMD~\cite{VAEHMD} and FLAG~\cite{FLAG} utilize Variational AutoEncoder (VAE)~\cite{VAE} and Normalizing flow~\cite{Flow}, respectively. Recent works~\cite{AGROL, BoDiffusion} leverage more powerful diffusion models~\cite{DDIM,DDPM} for motion generation, yielding promising results due to the powerful ability of diffusion models in modeling the conditional probabilistic distribution of full-body motion.

Contrasting with previous methods that model full-body motion in a comprehensive, unified framework, our approach acknowledges the complexities such methods impose on deep learning models, particularly in capturing the intricate kinematics of human motion. Hence, we propose a stratified approach that decouples the conventional full-body avatar reconstruction pipeline, first for the upper body and then for the lower body under the condition of the upper-body.

% Unlike all previous methods on this task that model full-body motion in a unified and comprehensive manner, we recognize the challenges this poses to deep learning models in fully grasping the complex kinematics of human motion. Hence, we propose disentangle the full body motion space, allowing deep learning models to concentrate on a narrower range of movements and interactions, thereby streamlining the reconstruction process.

\subsection{Human Motion Generation}
%Various conditions have been explored in human motion generation research, 
Human motion generation is explored under various input conditions, including text~\cite{T2M-GPT,TEMOS,motiondiffusion,MotionGPT,MotionClip}, action labels~\cite{Action2Motion,ActionVAE}, 3D scenes and objects~\cite{hassan21stochastic,xu2023interdiff,Pi_2023_ICCV}, and motion itself~\cite{chen2023humanmac,mao21generating,PoseGPT}. Our work shares similarities with two primary research streams. The first involves diffusion-based motion generation. For instance, \cite{motiondiffusion} is the first to utilize a diffusion model~\cite{DDIM,DDPM} for text-to-motion generation with a transformer network. \cite{Pi_2023_ICCV} develops a hierarchical generation pipeline for human-object interaction by generating initial keyframes in the motion sequence and then interpolating between them. Secondly, our approach parallels works like PoseGPT~\cite{PoseGPT}, MotionGPT~\cite{MotionGPT}, and T2M-GPT~\cite{T2M-GPT} in terms of representing human motions with discrete latent. These studies also utilize a VQ-VAE~\cite{VQ-VAE} to encode human motion into a discrete latent space, facilitating the subsequent generation process.

The task we aim to address significantly deviates from traditional motion generation tasks like text-to-motion generation, which typically aims to create motions that align with textual descriptions. Our goal is distinctly different: we focus on accurately reconstructing human motion using solely sparse observations.

% Our work shares similarities with PoseGPT~\cite{PoseGPT}, MotionGPT~\cite{MotionGPT}, and T2M-GPT~\cite{T2M-GPT}. These works also employ a VQ-VAE~\cite{VQ-VAE} to encode human motion into a discrete latent space, enabling the generation of new human motions based on past motion~\cite{PoseGPT} and textual descriptions~\cite{MotionGPT, T2M-GPT}, respectively. However, our approach differs in that we specifically disentangle the upper and lower body of the human, encoding the motions into separate discrete latent. This reduces the embedding space and simplifies the learning process. While ~\cite{PCT} also achieves disentanglement of the human sub-structure, they do so implicitly using sub-codebooks in the latent space. In our work, we explicitly disentangle the human sub-structure, resulting in a more controllable reconstruction process.

% Furthermore, instead of directly employing a network to predict latent based on given conditions, we utilize a powerful diffusion model to generate motion latent from sparse observations. This choice accounts for the uncertain nature of the task and enhances the generation process.

\begin{figure*}
    \centering
    \includegraphics[width=0.99\linewidth]{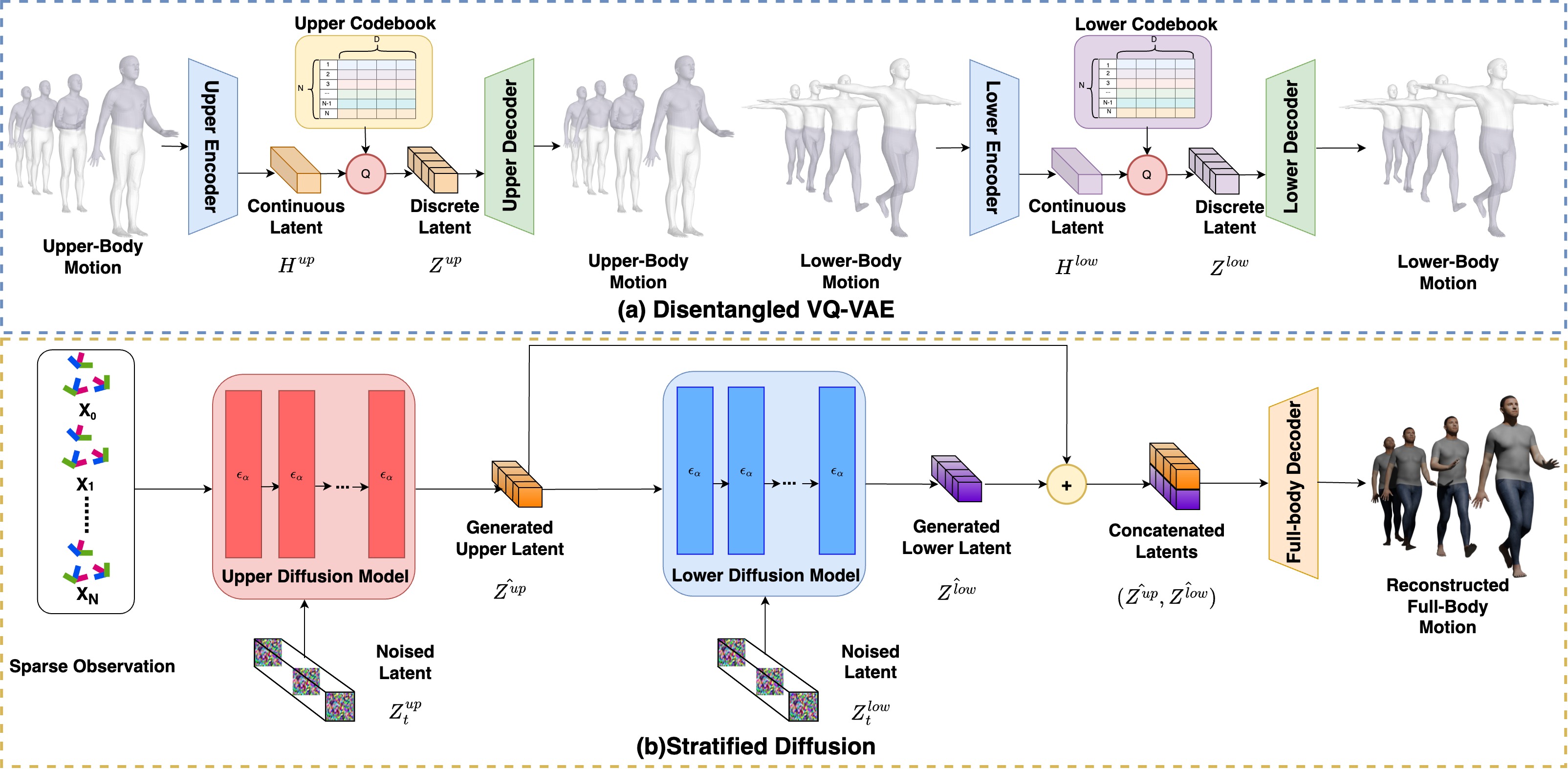}
    \caption{The overall architecture of our SAGE Net. It mainly contains two components: (a) Disentangled VQ-VAE for discrete human motion latent learning. To facilitate visualization, we incorporate zero rotations as padding for the lower body in the Upper VQ-VAE, and vice versa for the Lower VQ-VAE. Consequently, in the visualizations of the Upper VQ-VAE, the lower body remains in a stationary pose, whereas in the visualizations of the Lower VQ-VAE, the upper body is maintained in a T-pose. (b) The stratified diffusion model, which models the conditional distribution of the latent space for upper and lower motion. This model sequentially infers the upper and lower body latents, capturing the correlation between upper and lower motions. By employing a dedicated full-body decoder on the concatenated upper and lower latents, we can obtain full-body motion.}
    \label{fig:model}
\end{figure*}

\section{SAGE: Stratified Avatar Generation}
\label{sec:method}
This section introduces the proposed SAGE Network, following our observation about the connection relationship between upper-body and lower-body motions. The overall architecture of our SAGE Net is shown in~\cref{fig:model}. Disentangled latent representations for upper-body and lower-body motions are learned in \cref{fig:model} (a). Subsequently, as illustrated in \cref{fig:model} (b), we employ a stratified latent diffusion process for full body motion reconstruction.

% This process initiates with the generation of upper body motions, followed by the lower body motions, which are conditioned on the generated upper body motions.

%we first establish the 
%As summarized in \cref{fig:model}, we first establish the latent representation of stratified 3D avatar in \cref{fig:model} (a), and then we build a stratified latent diffusion for our end task in \cref{fig:model} (b). 

\subsection{Problem Statement and Notation}\label{subsec:problemformulation}
\paragraph{Input Signals.} Our paper follows the common setting of Head Mounted Devices (HMDs) inputs for motion generation, in which three sensors mounted on the head and, left and right hands are employed to perceive the corresponding joint motions. Formally, the raw input signals are denoted by a time-dependent vector function $\mathbf{X}(t) = (\mathbf{x}_h(t), \mathbf{x}_l(t), \mathbf{x}_r(t))$, where the subscripts $h$, $l$ and $r$ indicate the head, left hand, and right hand, and all these functions are with six degree of freedom for 3D rotation and translation under the global coordinate system. Joint rotations are represented by a six-axis representation, which has been demonstrated to be more suitable for network learning in previous works~\cite{zhou10continuity}.
Given a time interval with $T$ sampling points, the raw input signals can be denoted in a matrix $\mathbf{X}_{\rm raw} \in \mathbb{R}^{T\times (3\times (3+6))}$. To enhance the input signals, we follow \cite{AvatarPoser} to compute the positional velocities and angular velocities. This augmentation process adds a 9D input signal for each observed joint, resulting in an 18D input signal per joint at every timestamp. By combining these signals for all joints over all timestamps, we form the complete sparse input signal, represented as $\mathbf{X} \in \mathbb{R}^{T\times 54}$.

% \begin{figure}
%     \centering
%     \includegraphics[width=0.5\linewidth]{img/smpl_demo.pdf}
%     \caption{The demonstration of the kinematic tree in SMPL.}
%     \label{fig:smpl_demo}
%     \vspace{-5mm}
% \end{figure}

\paragraph{Kinematic Tree and SMPL Representation.} As shown in~\cref{fig:teaser} (b), SMPL~\cite{SMPL} represents a human pose by a standard skeletal rig, which is widely adopted by current motion generation works. A pose $\theta^j(t)$ represents the relative rotation of joint $j$ at $t$th frame with respect to its parent in the kinematic tree. The global rotations $G(\theta^j(t))$  can be calculated by:
\begin{equation}
    G(\theta^j(t)) = \prod_{k \in A(j)} \theta^k (t)
\end{equation}
%Here $\theta$ is represented as a rotation matrix and we omit the translation from 6-d axis angle to rotation matrix. 
where $A(j)$ denotes the ordered set of joint ancestors of joint $j$. 

Articulated motion representation based on the human kinematic tree, is key for realistically simulating human motions and enables efficient control over joint parameters. However, such an intricate articulated motion representation poses a significant challenge for models to learn effectively. In this work, we seek to disentangle this complex representation to enable the model to focus on a limited set of motions and interactions, thereby simplifying the learning process. 

Nevertheless, separating full-body human motions into distinct parts is nontrivial due to the complex correlations among joints. We revisit the human kinematic tree defined in SMPL model, where the upper and lower half-body is connected solely via a root joint. This insight from SMPL model provides a natural solution to separate the articulated full-body motion into two distinct parts: upper-body motion and lower-body motion. Notably, the root joint is included in both two parts as a central element since the parameters of all other joints in each half-body are defined in the local coordinate system of the root joint.

\paragraph{The Outputs.} As discussed in the last paragraph for SMPL representation~\cite{SMPL}, the problem of 3D body avatar generation comes down to the full-body motion estimation of 22 joints (including the root joint), denoted in the set function $\Theta(t) = \left\{\mathbf{\theta}^i(t) \in {\rm SE(3)} | t \in \{t_1,\ldots, t_T\}\right\}$ as the expected output of our problem. 
Based on the discussion of SMPL model with the disentangling nature of upper and lower body, we redefine the set function $\Theta(t)$ in the disentangled way by $\Theta(t) = \Theta_{\rm upper} (t) \cup \Theta_{\rm lower} (t)$, where $\Theta_{\rm upper} (t) = \{\theta^0(t),\ldots, \theta_{u}^{b_u}(t)\}$, $\Theta_{\rm lower} = \{\theta^0(t),\ldots, \theta_{l}^{b_l}(t)\}$. These two subsets have only one intersected joint: root joint $\theta^0$, and $b_u=13$ and $b_l=8$ denote the number of rest joints in the upper and lower body, respectively. For the final output of our method, the dimension of the underlying motion variables is $22\times 6 = 132$ at every timestamp.

\subsection{Disentangled Motion Representation}
\label{subsec:represetation}
In this section, our objective is to disentangle full-body human motions into upper-body and lower-body parts and encode them to discrete latent spaces. This can effectively reduce the complexity and burden of encoding since each encoding takes care of only half-body motions.

We employ two autoencoders, i.e., VQ-VAE~\cite{VQ-VAE, CVAE}, with identical architecture to learn the discrete latent spaces for upper-body and lower-body motions, respectively. As shown in~\cref{fig:model} (a), our VQ-VAE model consists of an encoder and a decoder. The encoder $E$ takes the motion sequence $\Theta = \{\theta_i\}_{i=1}^T$ as input and encodes it into a series of continuous latent $E(\Theta) = H$, where $H =\{h_i\}_{i=1}^{T/l}$, and $l$ is the temporal down-sampling rate of input motion sequence.

To quantize the continuous latent, we define the discrete motion latent space by a codebook $C= \{c_j\}_{j=1}^N \in \mathbb{R}^{N \times D}$, where $N=512$ is the number of entries in the codebook and $D=384$ is the dimension of each element $c_j$. The operation $Q$ quantizes the continuous latent $h_i$ into discrete latent $z_i$ by finding its most similar element in $C$:
\begin{equation}
    {z_i = Q(h_i) = \underset{c_j\in C}{\arg\min}  \left \| h_i - c_j \right \|_2}
\end{equation}
Since continuous latent from all data samples share the same codebook $C$, all the real motions in the training set could be expressed by a finite number of bases in latent space.

Subsequently, the quantified latents $Z$ are fed into the decoder to reconstruct the original motions, given by $\hat{\Theta} = D(Z)$. The training process involves the joint optimization of the encoder and decoder by minimizing the following loss over the training dataset:
\begin{equation}
\small
    \begin{split}
        Loss_{vq} &= {\rm Smooth}_{L_1}(\hat{\Theta}, \Theta) + \left \| FK(\hat{\Theta}) -  FK(\Theta)\right\|_2 \\ &+ \left \| {\rm sg}[Z] - H\right\|_2 + \beta \left \| Z - {\rm sg}[H]\right\|_2
    \end{split}
\end{equation}
% \begin{multline}
%     Loss_{vq} = smooth_{L_1}(\hat{\Theta}, \Theta) + \left \| sg[Z] - Q(Z)\right\|_2 \\ + \beta \left \| Z - sg[Q(Z)]\right\|_2
% \end{multline}
Here, $\rm sg$ denotes the stop gradient operator, $FK$ stands for forward kinematic function and $\beta$ is a hyperparameter. We have two independent VQ-VAEs for upper-body and lower-body motion encoding, which we refer to as VQ-VAE$_{up}$ and VQ-VAE$_{low}$.

% The VQ-VAE, introduced in~\cite{VQ-VAE}, revolutionized generative models by 
% allowing the model to acquire discrete representations. 
% Since then, numerous motion generation works~\cite{PoseGPT, T2M-GPT, PCT}
% have leveraged the power of VQ-VAE for their respective methodologies.

% Our main difference from theirs lies in that we adopt an online approach.
% For the online approach, there are two options to consider. 
% The first option involves taking a single frame $y_t$ as the input and 
% producing the corresponding output $\hat{y_t}$. 
% The second option involves taking a sequence of frames $\{y_i\}_t^{t+L}$ 
% as the input and producing the output $\hat{y}_{t+L}$, where $L$ is the sequence
% length. The first option fails to capture the temporal information, which can 
% result in significant discontinuity or abrupt transitions between two consecutive 
% frames. Therefore, we opt for the second option, where the objective is to 
% reconstruct one frame $\hat{y_{t+L}}$ based on a single frame ${y}_{t+L}$
% while utilizing the sequence $\{y_i\}_t^{t+L-1}$ to retain temporal information 
% and maintain a temporally coherent latent space.

% \begin{figure}[t]
%     \centering
%     % \fbox{\rule{0pt}{2in} \rule{0.9\linewidth}{0pt}}
%      \includegraphics[width=1.0\linewidth]{img/vqvaepipeline3.png}
%      \caption{VQVAE pipeline
%      It is set in Roman so that mathematics (always set in Roman: $B \sin A = A \sin B$) may be included without an ugly clash.}
%      \label{fig:vqvae}
%   \end{figure}

\subsection{Stratified Motion Diffusion}
\label{subsec:cascadediffuison}
After encoding and expressing different human motions as latents, we aim to properly sample from the latent space for full-body motion reconstructions and match the sparse observations.
%Building upon the achievements of latent diffusion models in modeling conditional probabilistic distributions for various latent representations, such as images~\cite{LatentDiffusion}, videos~\cite{VideoDiffusion}, and 3D objects~\cite{3DDiffusion}, we adapt the diffusion model to generate upper-body and lower-body latents from sparse observations.

% In order to reconstruct the full body based on the disentangled latents, a straightforward approach is to utilize two diffusion models in parallel. These two models using the sparse observation as only conditioning and predict the upper latents and lower latents, respectively. The predicted latents are then fed into the decoders of $VQVAE_{upper}$ and $VQVAE_{lower}$ to decode the motion of upper and lower body, which are merged together to obtain the full-body motion.
Although disentangling the full-body motions into upper and lower parts enhances effectiveness and efficiency for motion representation learning, it's crucial to include the correlation between two body parts during generation. Otherwise, severe inconsistency would be witnessed in reconstructed full-body motions. To this end, we propose Stratified Motion Diffusion to sample upper-body and lower-body latent in a cascaded manner with explicit considerations of the correlations mentioned above.

%, which could be overlooked if upper and lower latents are generated separately from sparse observations. To address this, we propose a serial conditioning strategy to generate full-body motion from disentangled latents, capturing the interplay between the upper and lower body.

Since the sparse observations are all from the upper body (e.g., head and hand sensors), we first generate upper-body latent $\hat{z^{up}}$ by upper diffusion model conditioning on the sparse observations $X$.Thus the training objective of the upper diffusion model is:
\begin{equation}
    \begin{split}
    L_{up}:=\mathbb{E}_{z^{up}, \epsilon \sim \mathcal{N}(0,1), k}\left[\left\|\epsilon-\epsilon_{\alpha}\left(z^{up}_{k}, X, k\right)\right\|_{2}^{2}\right]
    \end{split}
\end{equation}
where $\epsilon$ is the random noise from the normal distribution, $\epsilon_{\alpha}$ is the noise predictor of the diffusion model with network parameters $\alpha$, and $k$ is the diffusion time step.

%Since the sparse observations are all from upper half-body (e.g., head and hand sensors),  the upper latents  is easier to learn compared with the lower ones as the sparse observation includes the ground truth tracking signals of the head and hands, which are all part of the upper body. 
Compared with upper-body latent prediction, directly predicting the lower-body latent $\hat{z^{low}}$ from the same sparse observations is more challenging due to the absence of direct tracking or supervision for any of the lower-body joints. To make the prediction more physically meaningful, as shown in~\cref{fig:model} (b), we take both the sparse observations $X$ and the generated upper-body latent $\hat{z^{up}}$ as conditions for lower-body latent prediction by lower diffusion model. This design considers the correlation between two half-body parts and allows more information to be involved for the lower-body inference. The objective for lower diffusion model training is as follows:
\begin{equation}
    \footnotesize
    \begin{split}
    L_{low}:=\mathbb{E}_{z^{low}, \epsilon \sim \mathcal{N}(0,1), k}\left[\left\|\epsilon-\epsilon_{\alpha}\left(z^{low}_{k}, (X, \hat{z^{up}}),k\right)\right\|_{2}^{2}\right]
    \end{split}
\end{equation}

Once two half-body latent $\hat{z^{up}}$ and $\hat{z^{low}}$ are obtained, the full-body motions can be recovered with a decoder $\hat{\Theta} = D_{full}(\hat{z^{up}},\hat{z^{low}})$. Instead of directly using pre-trained upper and lower decoders in~\cref{fig:model} (a) to recover the corresponding half-body motions, we train this full-body decoder $E_{full}$ from scratch together with our stratified motion diffusion, which is further optimized to capture the correlations between half-body motions.  
%Upon deriving the estimated upper and lower latents, $\hat{z^{up}}$ and $\hat{z^{low}}$, we employ a specialized full-body decoder $E_{full}$ that takes $\hat{z^{up}}$ and $\hat{z^{low}}$ as input and decode the full-body motion: $\hat{\Theta} = E_{full}(\hat{z^{up}},\hat{z^{low}})$.
%Diverging from conventional latent diffusion model~\cite{LatentDiffusion} that use the decoders of VQ-VAE, the full body encoder here fuse the information from the upper and lower body for full body motion reconstruction, improving the performance.

% \textcolor{red}{why full body encoder should be further illustrated}.

% After obtaining the estimated upper and lower latents, $\hat{z^{up}}$ and $\hat{z^{low}}$, we diverge from previous approaches that utilize the decoders of VQVAE. Instead, we train a tailored full body decoder, denoted as $E_{full}$, which takes the generated disentangled latents as input. This full body decoder is designed to fuse the information from both the generated upper and lower body latents and reconstruct the full body motion. Therefore, the reconstructed full body motion can be obtained as $\hat{\Theta} = E_{full}(\hat{z^{up}},\hat{z^{low}})$. This full body decoder realizes the information interaction between upper and lower motion for full body motion reconstruction, further improves the performance of our SAGE Net. 

\subsection{Implementation Details}
\label{subsec:details}
Since both sparse observations and human motion occur sequentially, we utilize the widely adopted sequential network, i.e., transformer~\cite{transformer}, as the backbone network for the encoder and decoder in the disentangled VQ-VAE~\cite{VQ-VAE}, and the denoise network in the stratified diffusion model. We set temporal down-sampling rate $l=2$ to balance the computational cost and the performance. In our transformer-based model for upper-body and lower-body diffusion, we integrate an additional DiT block as described in~\cite{DiT}. During the training of the latent diffusion model, instead of predicting noise $\epsilon_k$ as formulated by the standard latent diffusion model~\cite{LatentDiffusion}, we follow~\cite{ramesh22hierarchical,motiondiffusion} and directly predict the latent $z$ itself, as we find that this operation can significantly reduce the sampled time steps during inference stage. For training 
decoders, i.e., $E_{up}$, $E_{low}$ and $E_{full}$, in addition to the rotation-level reconstruction loss, we incorporate the forward kinematic loss proposed in~\cite{AvatarPoser} and the hand loss described in~\cite{AvatarJLM}.

For the inference stage, we evaluate our model in an online manner. Specifically, we fix the sequence length at 20 for both the input and the output of our model, and only the last pose in the output motion sequence is retained. Given a sparse observation sequence, we apply our model using a sliding window approach. For the first 20 poses in the motion sequence, we predict by padding the sparse observation sequence $x$ at the beginning with the first available observation. We make this choice considering the practicality and relevance of online inference in real-world application scenarios. This allows the motion sequences to be predicted in a frame-by-frame manner. 

In addition, we employ a simple two-layer GRU~\cite{GRU} on the top of the full body decoder as a temporal memory to smooth the prediction of the output sequence with minimal computational expense, and we term it as a Refiner. To train this Refiner, we use the same velocity loss as ~\cite{AvatarJLM}. Our model takes 0.74ms to infer 1 frame on a single NVIDIA RTX3090 GPU.

\section{Experiments and Evaluation Metrics}
\begin{table*}[h!]
  \centering
  \begin{tabular}{@{}lcccccccccc@{}}
    \toprule
     Method &MPJRE & MPJPE & MPJVE & Hand PE & Upper PE & Lower PE & Root PE  & Jitter\\
    \midrule
    Final IK~\cite{FinalIK} &16.77 & 18.09 &59.24 & - & - & - & -  & - \\
    LoBSTr~\cite{LoBSTr} &10.69 & 9.02 & 44.97 & - & - & - & - & -  \\
    VAR-HMD~\cite{VAEHMD} &4.11 & 6.83 & 37.99 & - & - & - & - &-  \\
    Avatarposer~\cite{AvatarPoser} & 3.08 &  4.18 & 27.70 & 2.12 &1.81 & 7.59 & 3.34  & 14.49\\
    AvatarJLM~\cite{AvatarJLM} & 2.90 & 3.35 & 20.79 & 1.24& 1.42 & 6.14 & 2.94  & 8.39\\
    AGRoL (Online)~\cite{AGROL} & 2.96 & 4.26 & 79.12 & 1.51 & 1.73 & 7.91 & 3.78  & 84.79\\ 
    AGRoL (Offline)~\cite{AGROL} & 2.66 & 3.71 & 18.59 & 1.31 & 1.55 & 6.84 & 3.36 & 7.26 \\
    % Ours(down2) & 2.50  & 3.33 & 22.33 & 1.69 & 1.50 & 5.97 &\textbf{2.91} & 9.12 & 2,253.6 & 109.5\\
    % Ours(up4) & \textbf{2.48}  & \textbf{3.30} & 21.67 & 1.56 & \textbf{1.46} & \textbf{5.96} &\textbf{2.91} & 9.01 & 371.1\\
    %Ours & 2.54 & 3.26 & 21.99 & 1.36 & 1.42 &5.93 & 2.91 & 9.29 \\
    \rowcolor[RGB]{241, 241, 255} 
    Ours & 2.53 & 3.28 & 20.62 & 1.18 & 1.39 & 6.01 & 2.95  & 6.55 \\
    \bottomrule
  \end{tabular}
  \caption{Evaluation results under setting S1.}
  \label{tab:result_p1}
\end{table*}

\begin{table}[h!]
\centering
\begin{tabular}{@{}l|ccc@{}}
\toprule
Method      & \multicolumn{1}{l}{MPJRE} & \multicolumn{1}{l}{MPJPE} & \multicolumn{1}{l}{MPJVE}  \\ \midrule
Final IK~\cite{FinalIK}    & 12.39 & 9.54 & 36.73 \\
CoolMoves~\cite{CoolMoves}   & 4.58  & 5.55 & 65.28 \\
LoBSTr~\cite{LoBSTr}      & 8.09  & 5.56 & 30.12 \\
VAE-HMD~\cite{VAEHMD}     & 3.12  & 3.51 & 28.23  \\
AvatarPoser~\cite{AvatarPoser} & 2.59  & 2.61 & 22.16 \\
AvatarJLM~\cite{AvatarJLM}   & 2.40  & 2.09 & 17.82 \\
AGRoL~\cite{AGROL}       & 2.25     & 2.17    & 16.26 \\
\rowcolor[RGB]{241, 241, 255} 
Ours        & 2.10  & 1.88  & 14.79  \\ 
\bottomrule
\end{tabular}
\caption{Evaluation results under setting S1 with the root joint as an additional input.}
\label{tab:result_p1_4input}
\end{table}

\begin{table}[h!]
\begin{tabular}{@{}l|cccc@{}}
\toprule
Method      & \multicolumn{1}{l}{MPJRE} & \multicolumn{1}{l}{MPJPE} & \multicolumn{1}{l}{MPJVE} & \multicolumn{1}{l}{Jitter} \\ \midrule
$\text{VAEHMD}^{\dagger}$~\cite{VAEHMD}      & -                         & 7.45                      & -                         & -                          \\
$\text{Humor}^{\dagger}$~\cite{Humor}      & -                         & 5.50                      & -                         & -                          \\
$\text{FLAG}^{\dagger}$~\cite{FLAG}        & -                         & 4.96                      & -                         & -                          \\ \midrule
AvatarPoser~\cite{AvatarPoser} & 4.70                      & 6.38                    & 34.05                     & 10.21                      \\
AGRoL~\cite{AGROL}       & 4.30                      & 6.17                      & 24.40                     & 8.32                       \\
AvatarJLM~\cite{AvatarJLM}   & 4.30                      & 4.93                      & 26.17                     & 7.19                       \\
Ours        & 4.62             & 5.86                             & 33.54                     & 7.13                          \\ \bottomrule
\end{tabular}
\caption{Evaluation result under setting S2. $\dagger$ indicates that these methods use additional inputs of pelvis location and rotation for training and inference, which are not directly comparable methods. The results of AvatarPoser~\cite{AvatarPoser} is provided by~\cite{AGROL}.}
\label{tab:result_p2}
\end{table}

\begin{table*}[h!]
  \centering
  \begin{tabular}{@{}lcccccccccc@{}}
    \toprule
     Method &MPJRE & MPJPE & MPJVE & Hand PE& Upper PE & Lower PE & Root PE & Jitter\\
    \midrule
    AGRoL(Online)~\cite{AGROL} & 3.09 & 4.31 & 109.29 & 1.79 & 1.80 & 7.95 & 3.86 & 121.78  \\
    AGRoL(Offline)~\cite{AGROL} & 2.83 & 3.80 & 17.76 & 1.62 & 1.66 & 6.90 & 3.53 & 10.08  \\
    AvatarJLM~\cite{AvatarJLM} & 3.14 & 3.39 & 15.75 & 0.69 & 1.48 & 6.13 & 3.04 & 5.33\\
    Avatarposer~\cite{AvatarPoser} & 2.72 & 3.37 & 21.00 & 2.12 & 1.63 & 5.87 & 2.90 & 10.24 \\
    \rowcolor[RGB]{241, 241, 255} 
    Ours & 2.41 & 2.95 & 16.94 & 1.15 &1.28 & 5.37 & 2.74 & 5.27 \\
    % Avatarposer & 4.44 & 5.91 & 31.79 & 16.20 & 5.48 & 6.53 & 3.74 & 13.80 \\
    % AGRoL(Online) & - & - & - & - & - & - & - \\
    % AvatarJLM & 3.18 & 3.40 & 17.16 & 0.66 & 1.50 & 6.14 & 3.08 & 7.47\\
    % Ours & \textbf{2.46} & \textbf{3.06} & 18.22 & 1.64& \textbf{1.42} & \textbf{5.42} & \textbf{2.77} & 6.81 \\
    % \midrule
    
    % Avatarposer & 4.42 & 5.83 & 31.33 & 15.89 & 5.26 & 5.37 & 3.69 & 13.53 \\
    % AGRoL(Online) & - & - & - & - & - & - & - \\
    % AvatarJLM & 3.09 & 3.26 & 16.74 & 0.64 & 1.43 & 5.89 & 2.86 & 7.14\\
    % Ours & \textbf{2.39} & \textbf{3.01} & 18.38 & 1.35 & \textbf{1.31} & \textbf{5.47} & \textbf{2.69} & 6.86 \\
    \bottomrule
  \end{tabular}
  \caption{Evaluation results under setting S3.}
  \vspace{-3mm}
  \label{tab:result_p3}
\end{table*}

\begin{figure*}[htb!]
    \centering
    \begin{minipage}[t]{0.19\linewidth}
    \centering
        \includegraphics[width=\linewidth]{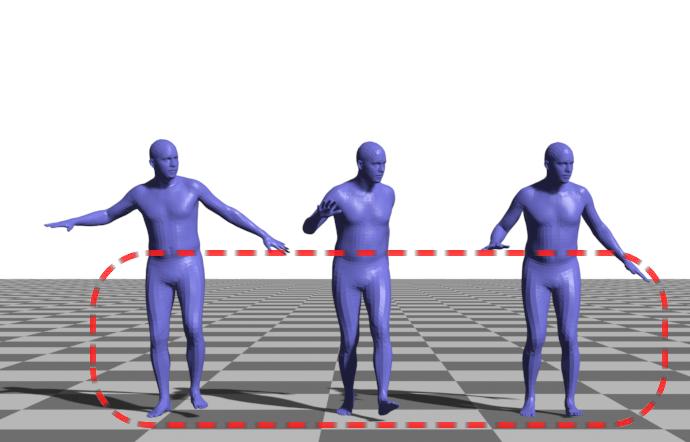} \\
         \includegraphics[width=\linewidth]{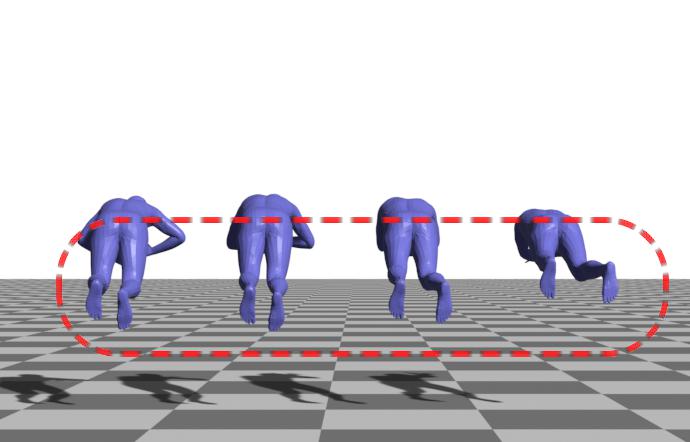} \\
        \includegraphics[width=\linewidth]{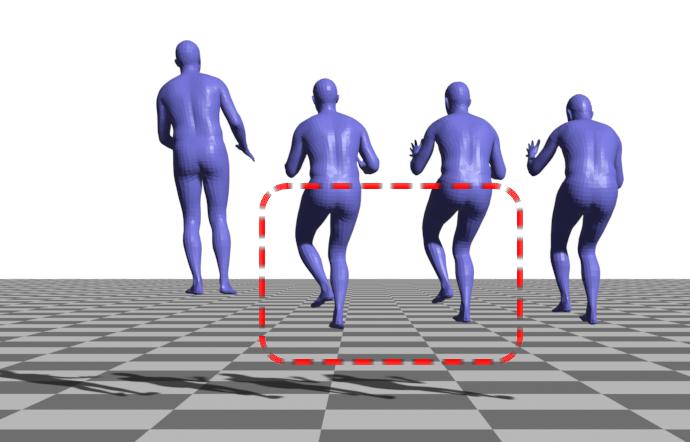} \\
        \includegraphics[width=\linewidth]{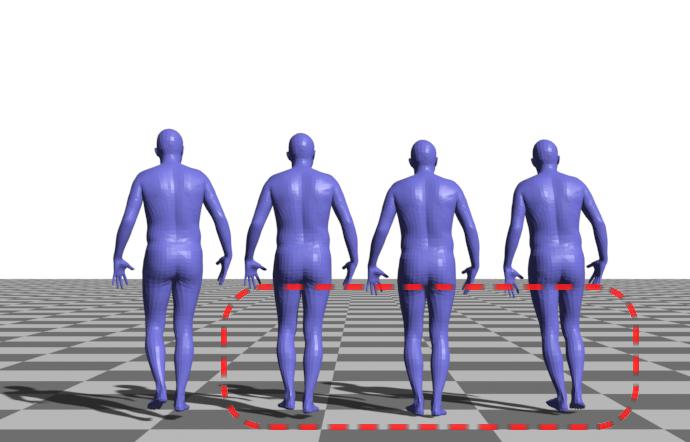}\\
        \includegraphics[width=\linewidth]{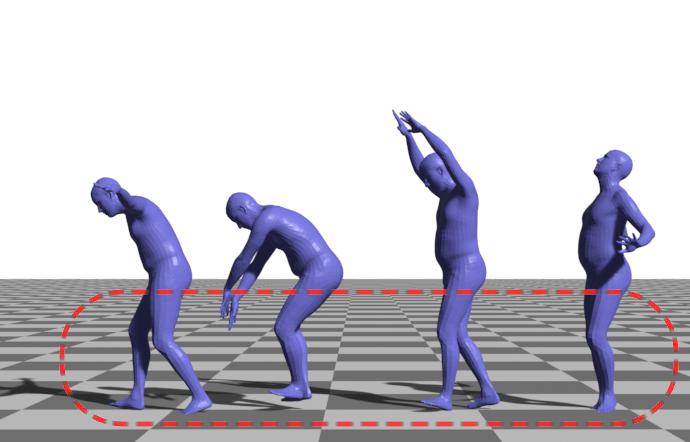} \\
        AvatarPoser~\cite{AvatarPoser}
    \end{minipage}
    \begin{minipage}[t]{0.19\linewidth}
    \centering      
        \includegraphics[width=\linewidth]{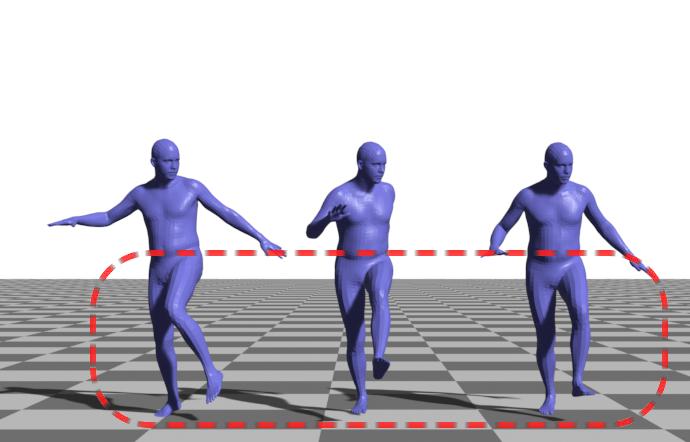} \\
         \includegraphics[width=\linewidth]{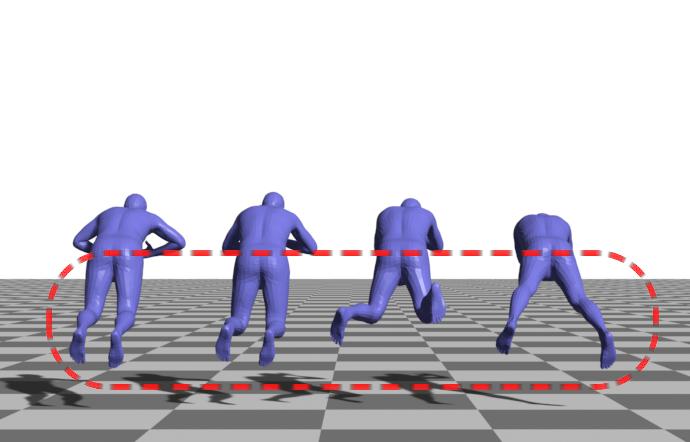} \\
        \includegraphics[width=\linewidth]{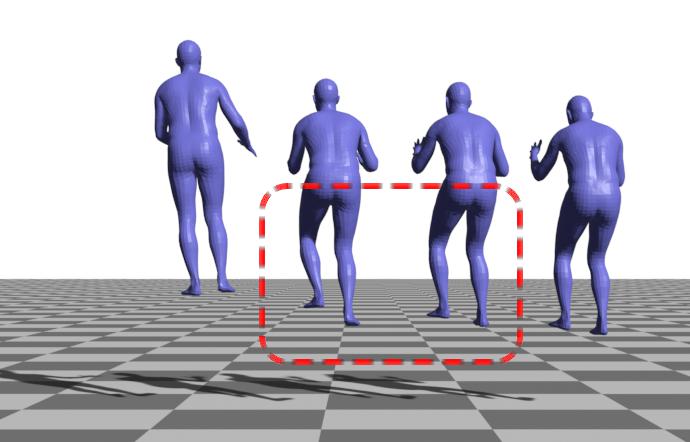} \\
        \includegraphics[width=\linewidth]{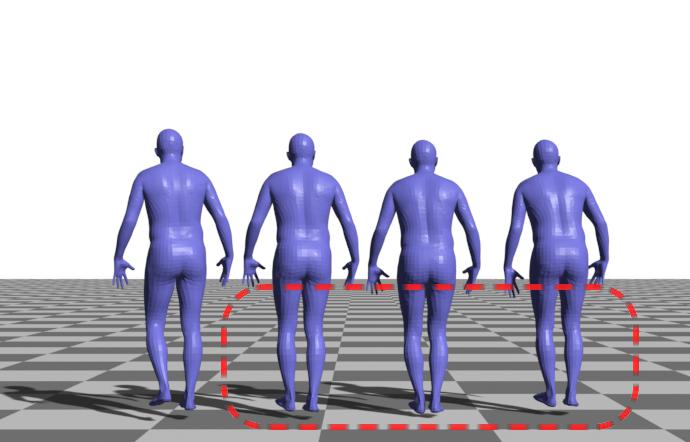}\\
        \includegraphics[width=\linewidth]{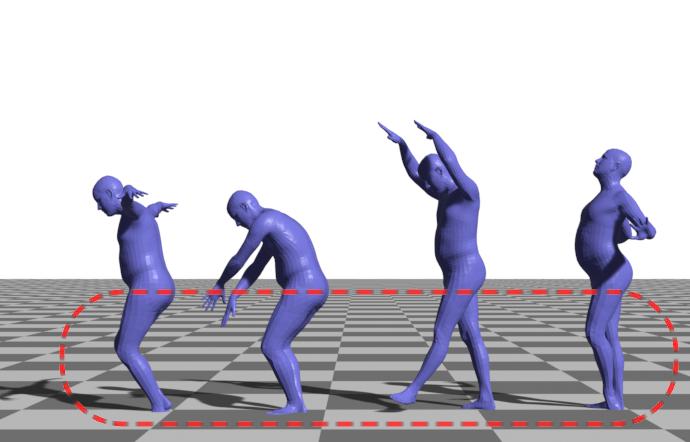} \\
        AGRoL~\cite{AGROL}
    \end{minipage}
    \begin{minipage}[t]{0.19\linewidth}
    \centering
        \includegraphics[width=\linewidth]{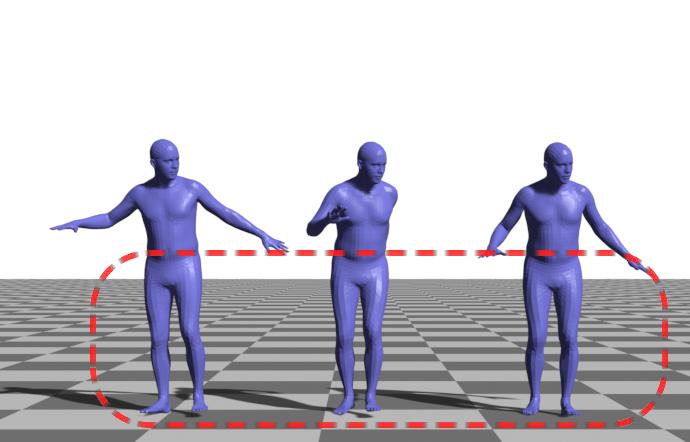} \\
         \includegraphics[width=\linewidth]{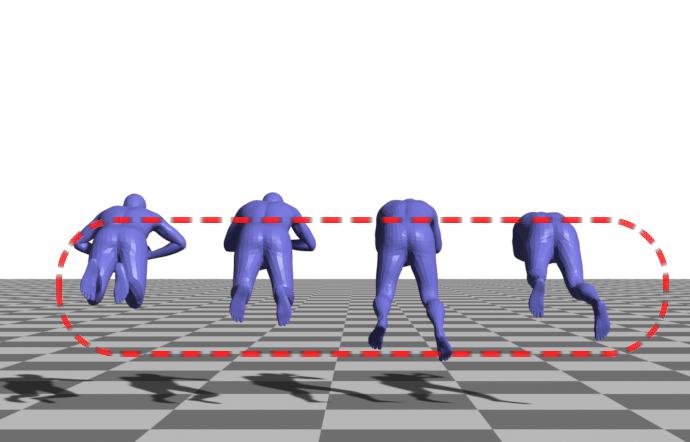} \\
        \includegraphics[width=\linewidth]{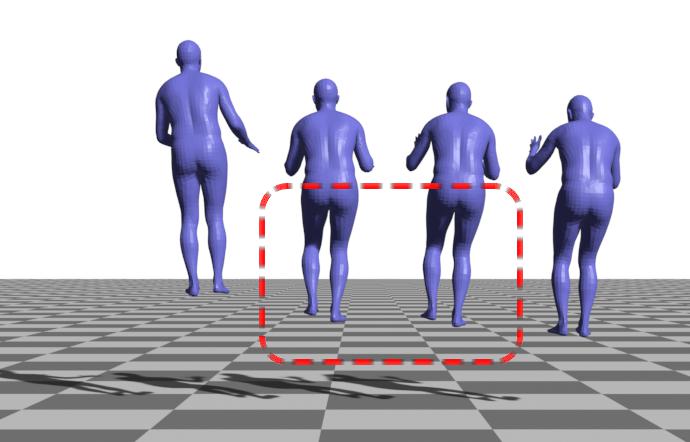} \\
        \includegraphics[width=\linewidth]{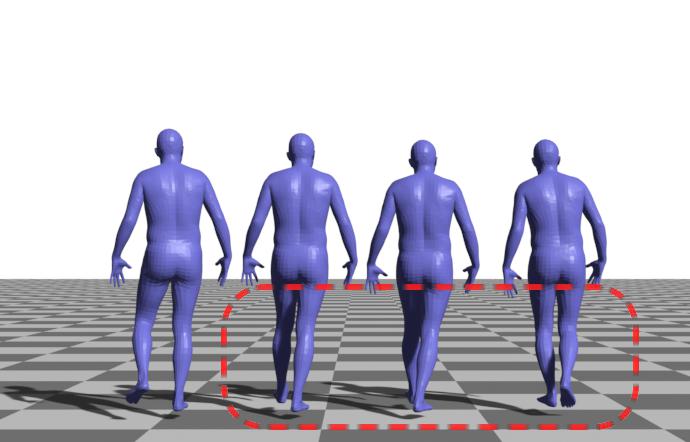}\\
        \includegraphics[width=\linewidth]{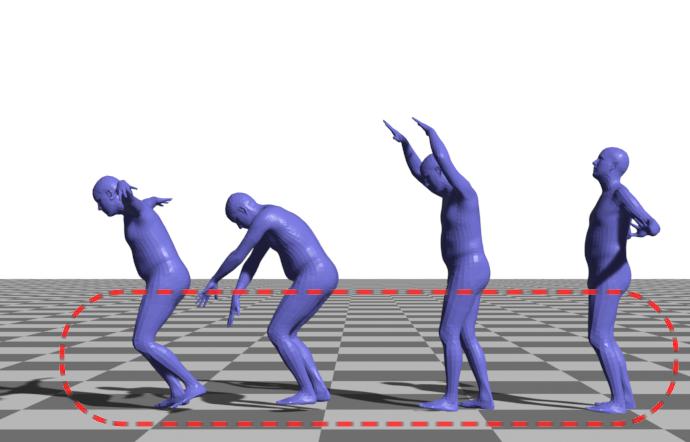} \\
        AvatarJLM~\cite{AvatarJLM}
    \end{minipage}
    \begin{minipage}[t]{0.19\linewidth}
    \centering
        \includegraphics[width=\linewidth]{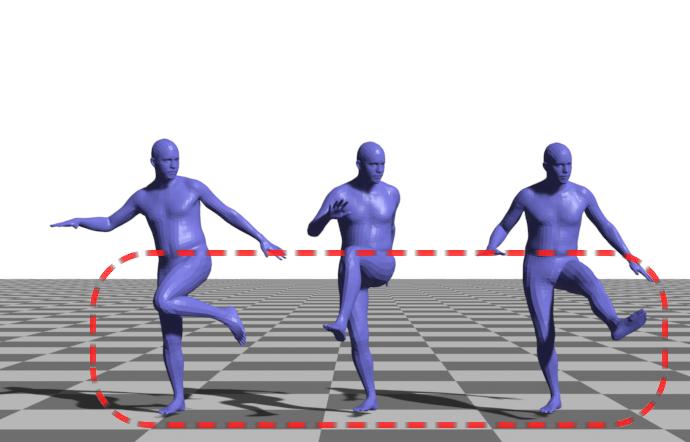} \\
         \includegraphics[width=\linewidth]{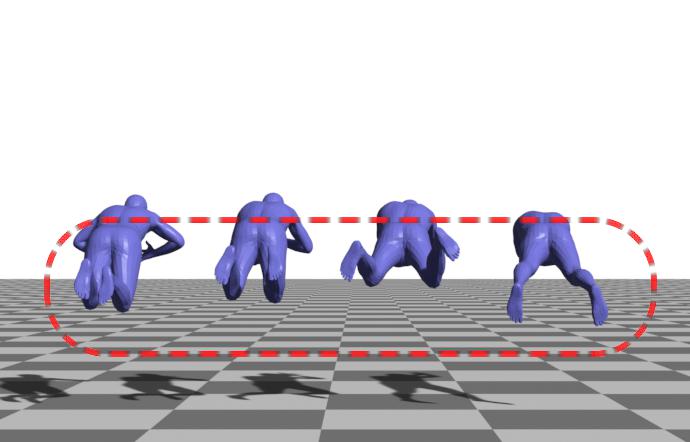} \\
        \includegraphics[width=\linewidth]{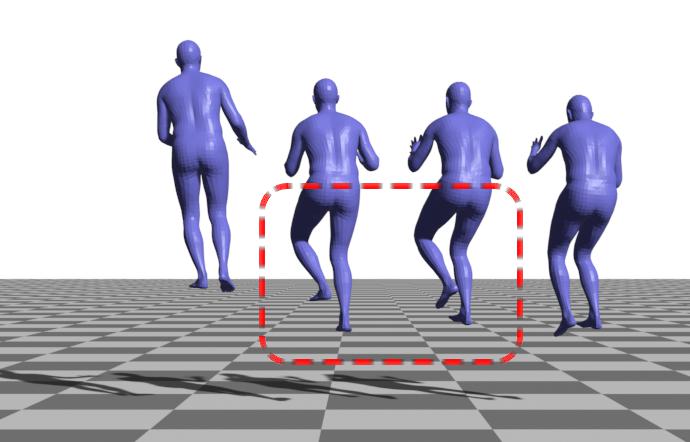} \\
        \includegraphics[width=\linewidth]{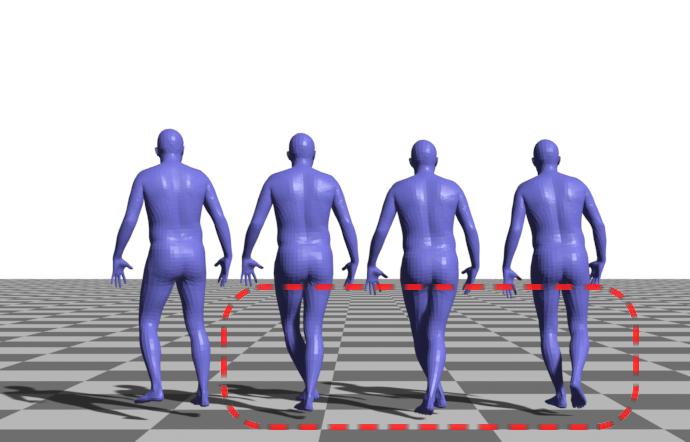}\\
        \includegraphics[width=\linewidth]{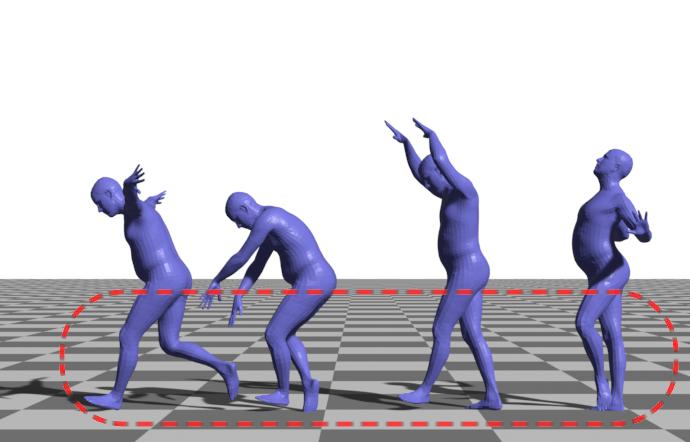} \\
        Ours
    \end{minipage}
    \begin{minipage}[t]{0.19\linewidth}
    \centering
        \includegraphics[width=\linewidth]{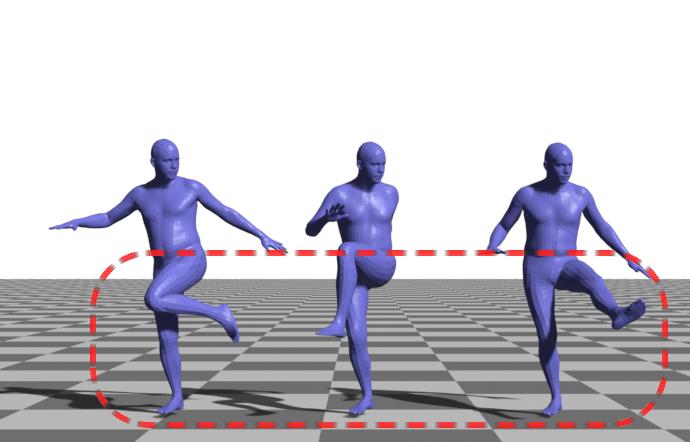} \\
         \includegraphics[width=\linewidth]{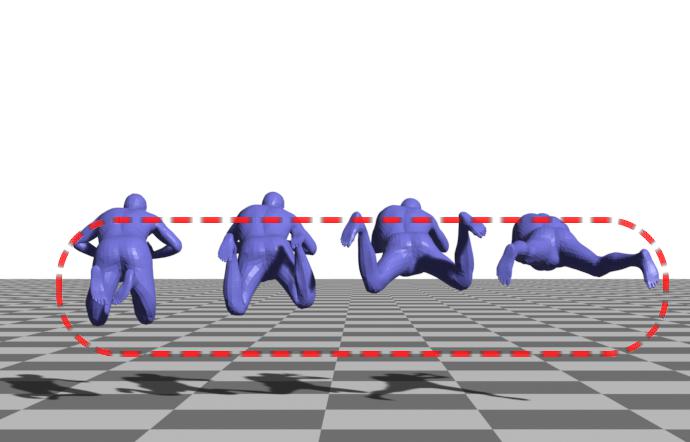} \\
        \includegraphics[width=\linewidth]{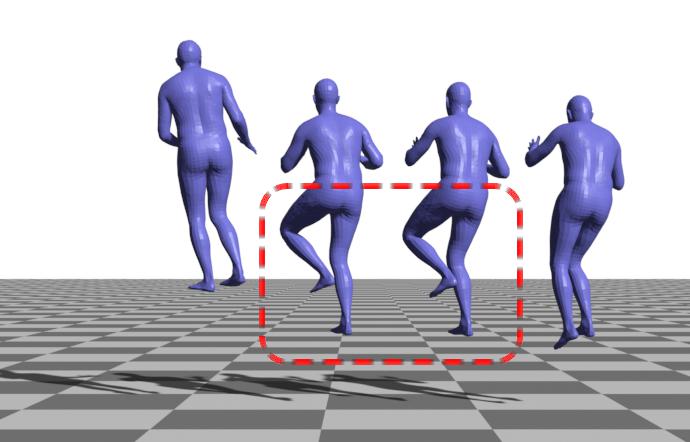} \\ 
        \includegraphics[width=\linewidth]{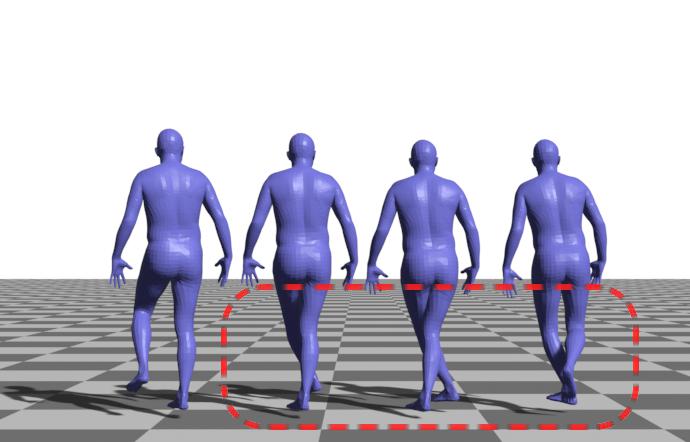}\\
        \includegraphics[width=\linewidth]{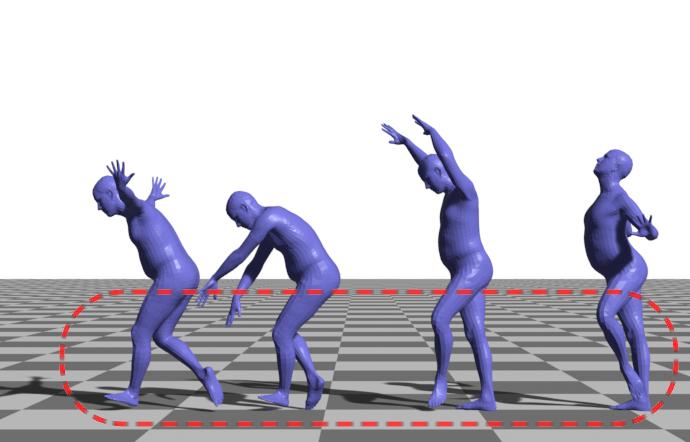}\\
        GT
    \end{minipage}
    \caption{Visualization results compared with other methods. All models are trained under setting S1.}
    \label{fig:vis}
\end{figure*}

\begin{figure}[h!]
    \centering
    \begin{minipage}[t]{0.3\linewidth}
    \centering
        \includegraphics[width=\linewidth]{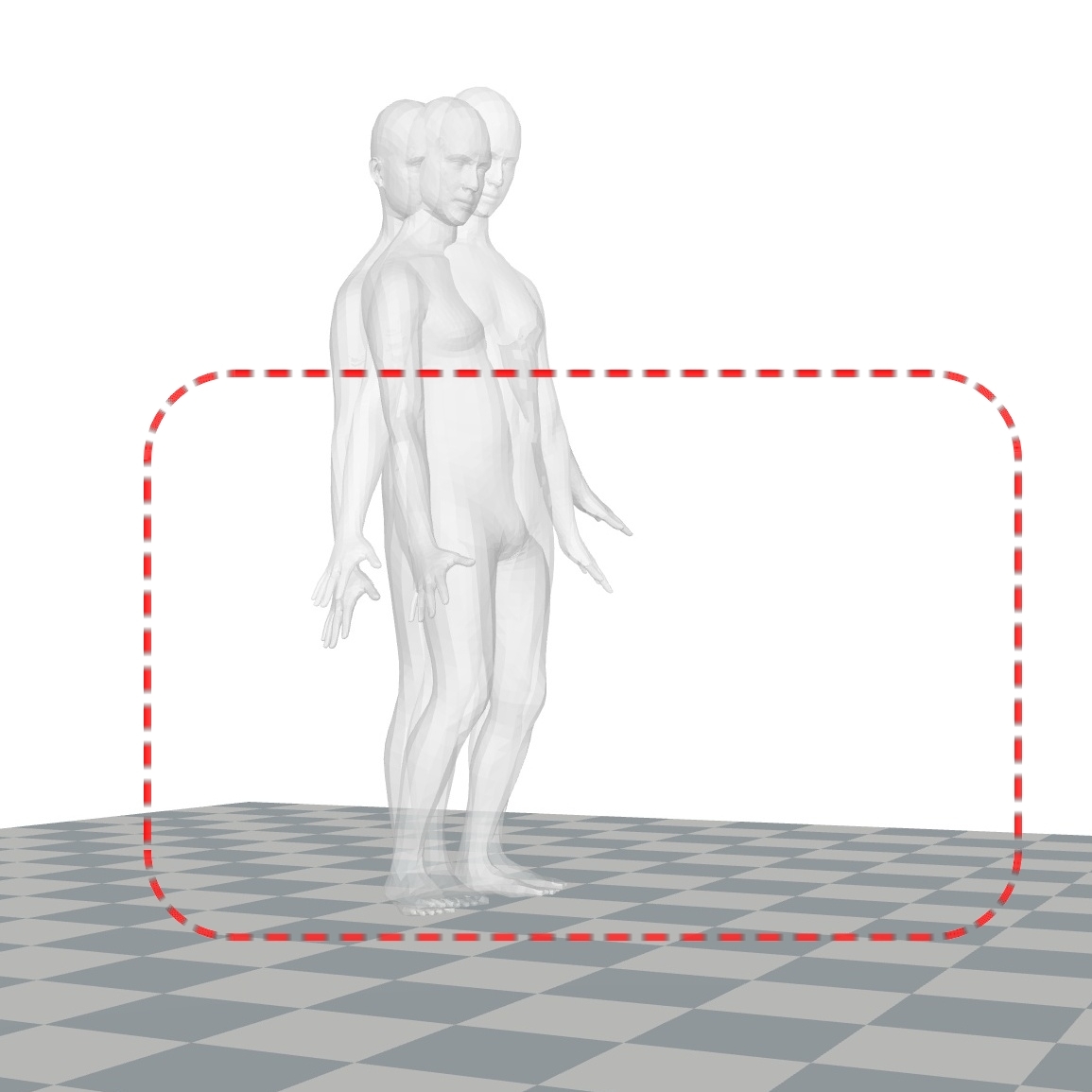} \\
        \includegraphics[width=\linewidth]{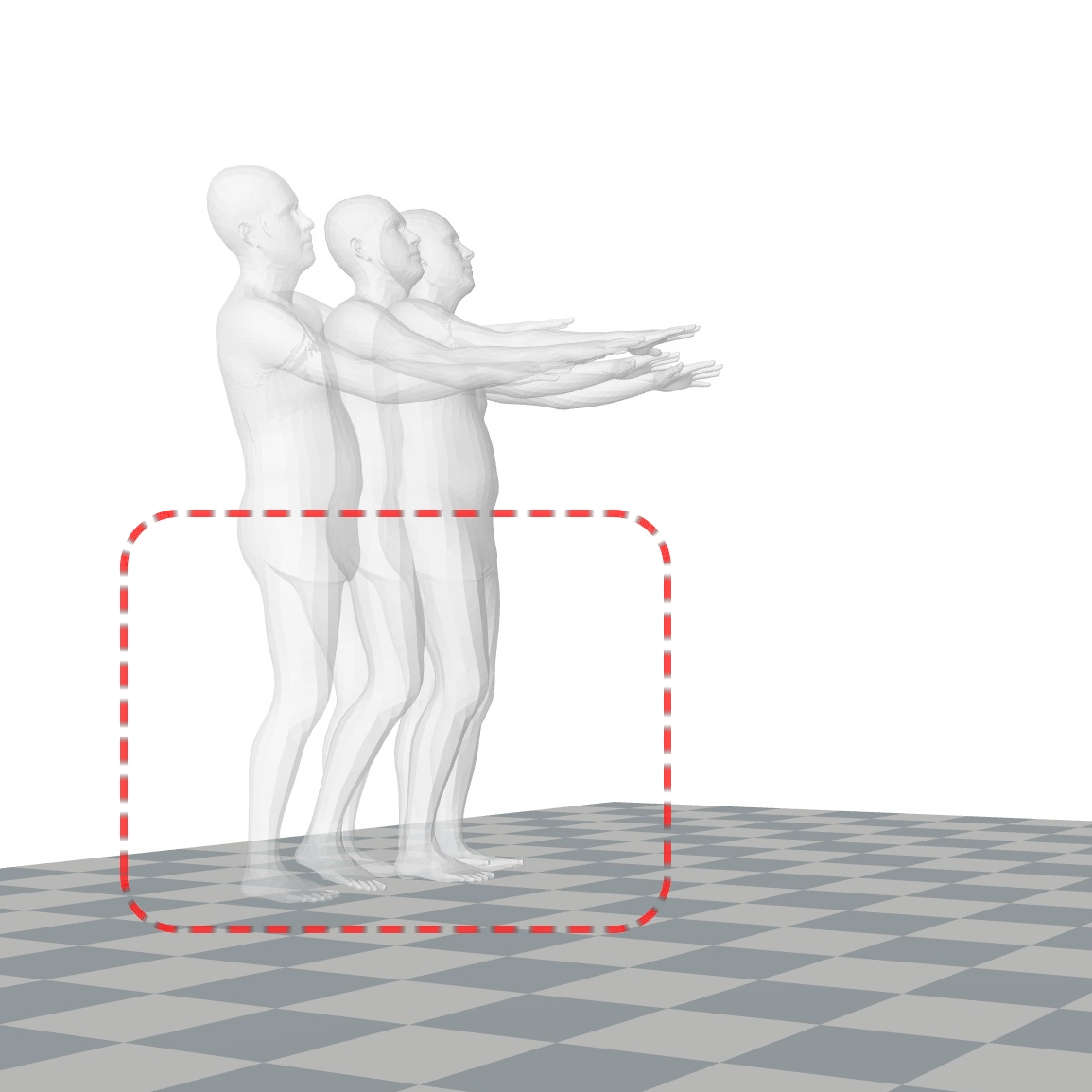} \\
        AvatarJLM~\cite{AvatarJLM}
    \end{minipage}
    \begin{minipage}[t]{0.3\linewidth}
    \centering
        \includegraphics[width=\linewidth]{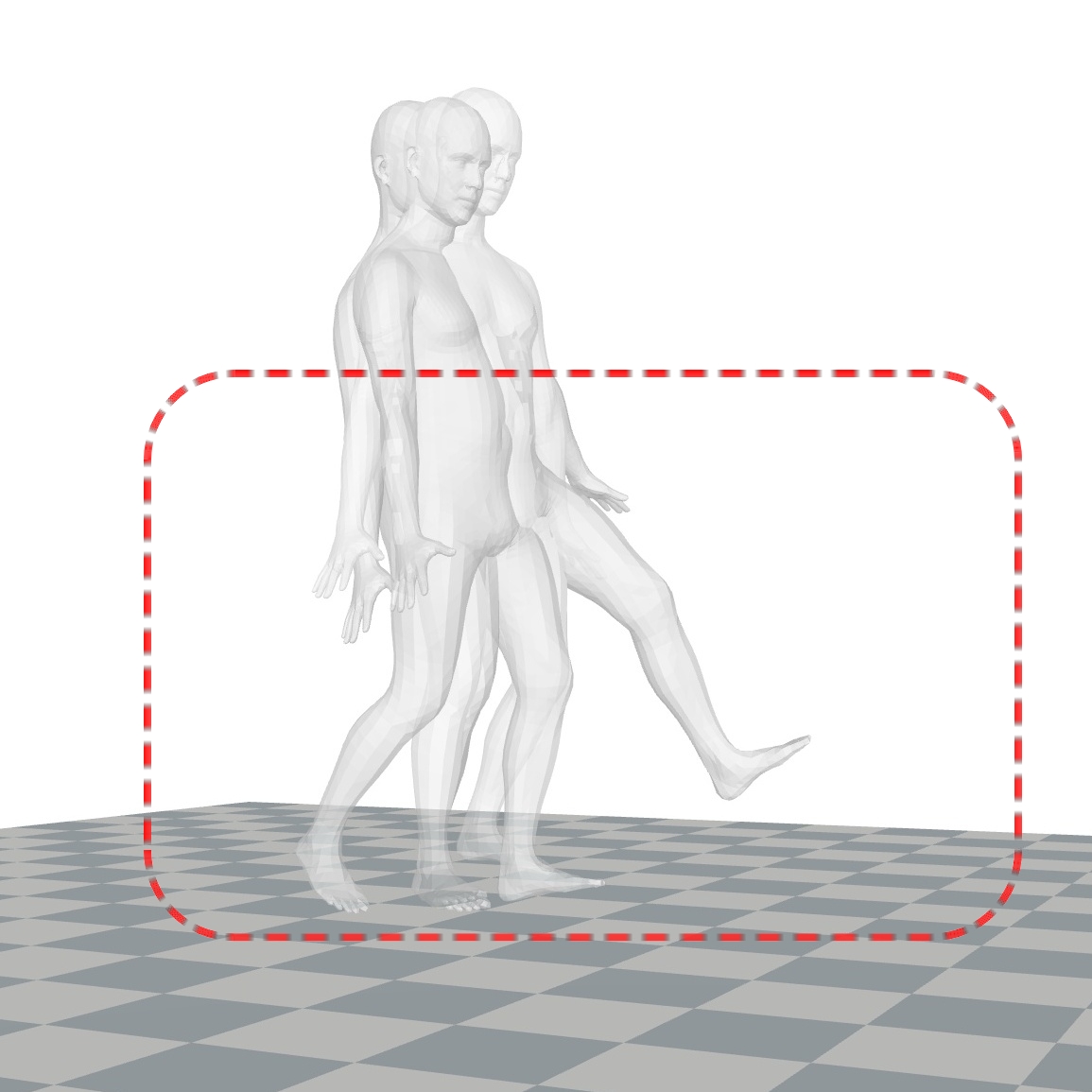} \\
        \includegraphics[width=\linewidth]{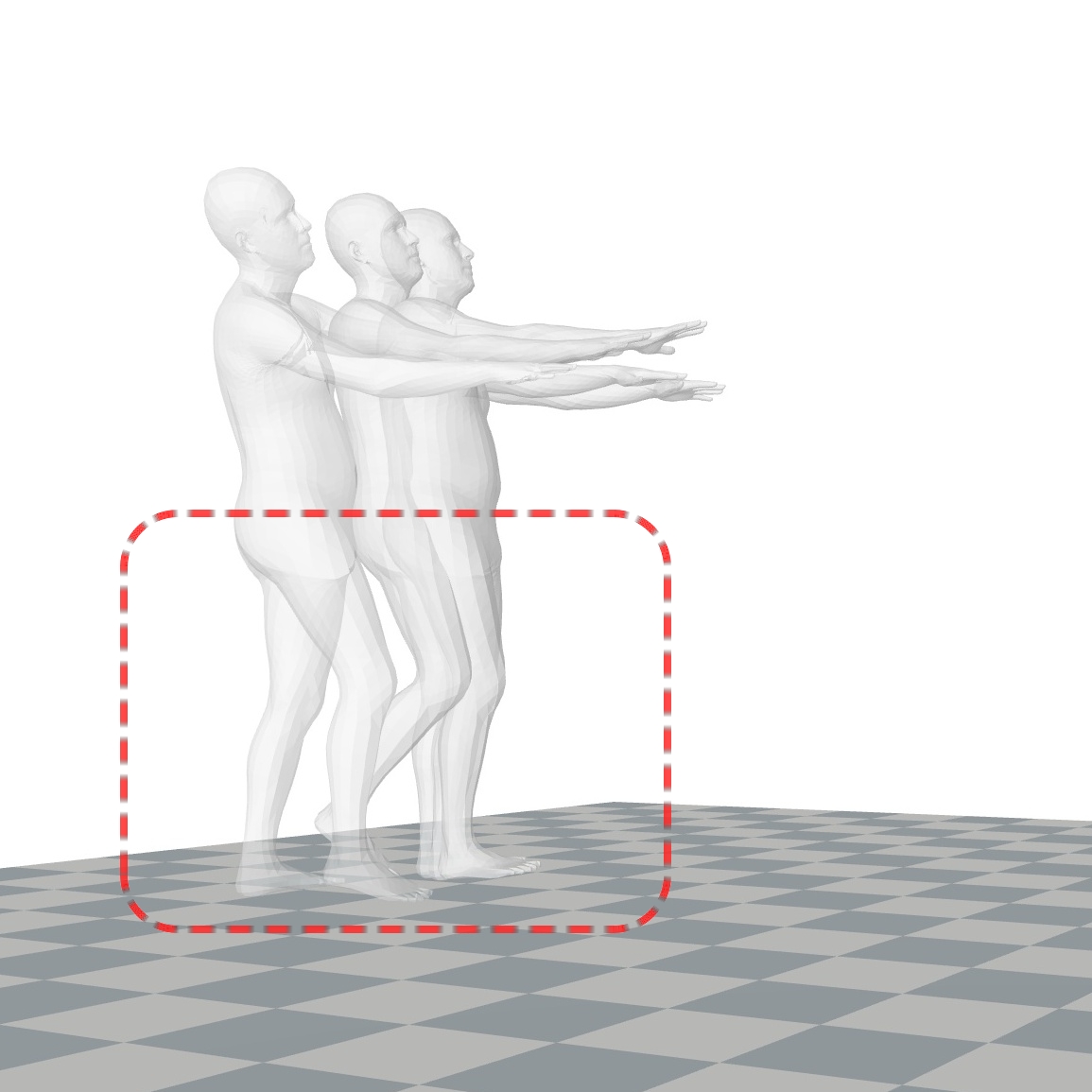} \\
        Ours
    \end{minipage}
    \begin{minipage}[t]{0.3\linewidth}
    \centering
        \includegraphics[width=\linewidth]{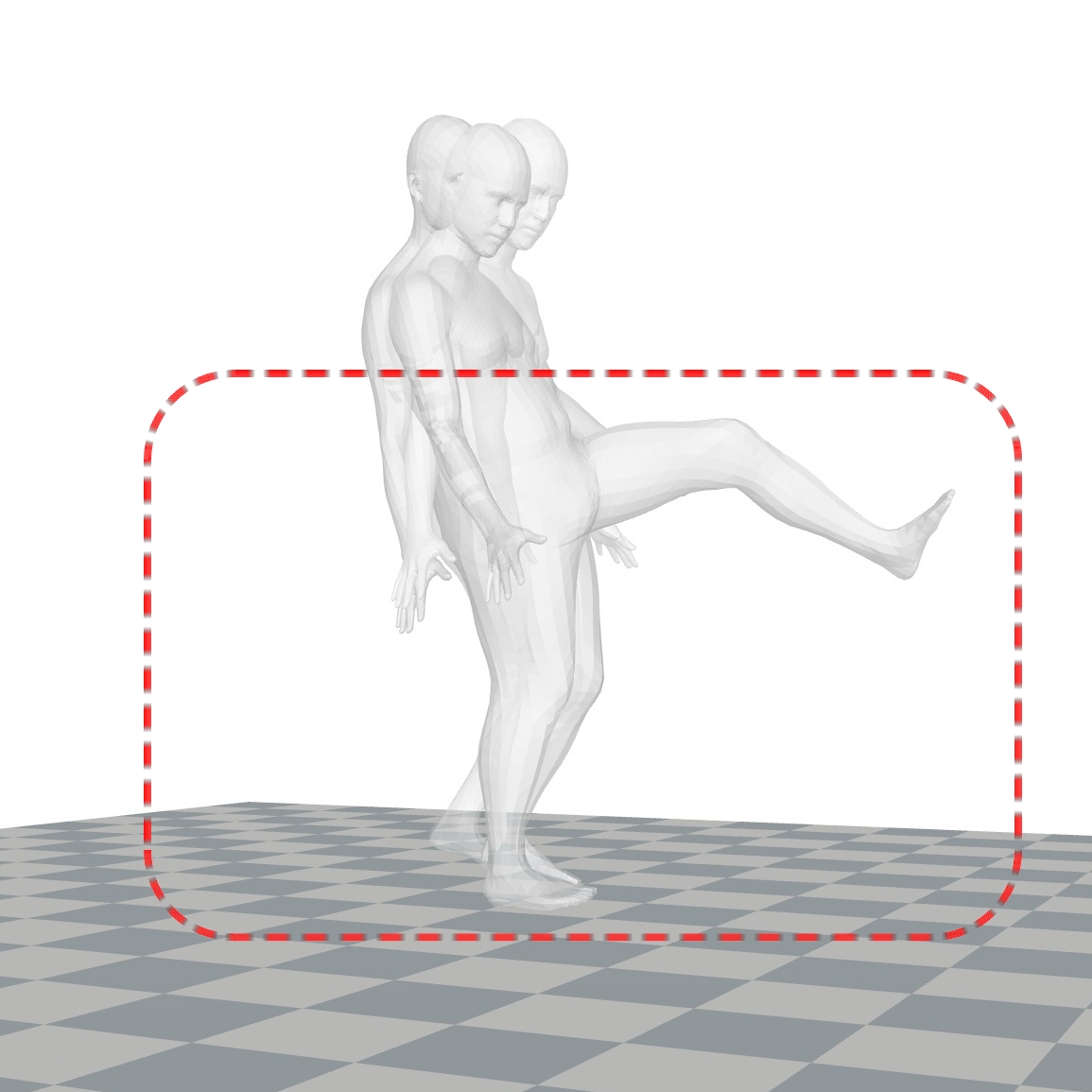} \\
        \includegraphics[width=\linewidth]{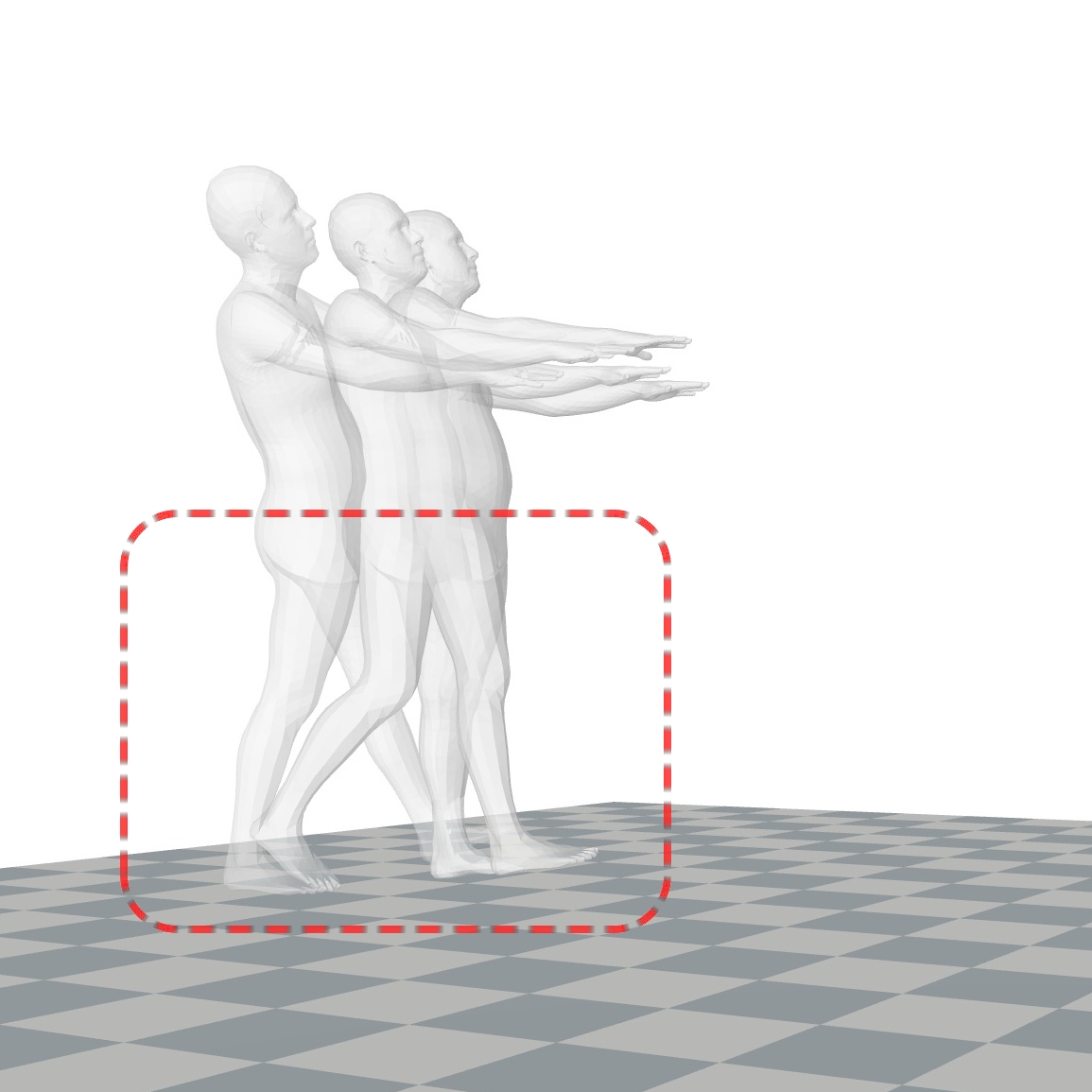} \\
        GT
    \end{minipage}
    \caption{Visualization results on real data.}
    \label{fig:vis_real}
\end{figure}

\begin{table}[h!]
  \centering
   \resizebox{0.97\linewidth}{!}{
  \begin{tabular}{@{}lcccccc@{}}
    \toprule
     Method &MPJRE & MPJPE & MPJVE & Jitter\\
    \midrule
    w/o Disentangle &2.64 & 3.62 &33.18 &25.07 \\
    w/o Full-Body Decoder & 2.71 &3.69 & 26.07 & 10.80\\
    %w/o Refiner & 2.54 & 3.26 & 21.99  &9.29 \\
    5 Part Disentanglement & 2.63 & 3.48 & 20.32 & 7.16\\
    \rowcolor[RGB]{241, 241, 255} 
    Ours & 2.53 &3.28 & 20.62 & 6.55  \\
    \bottomrule
  \end{tabular}
  }
  \vspace{-1mm}
  \caption{Ablation results of different components in SAGE Net under setting S1.}
  \label{tab:ablation_disentangle1}
  \vspace{-2mm}
\end{table}

\begin{table}[h!]
  \centering
  \begin{tabular}{@{}ccccccccccc@{}}
    \toprule
     Method & Lowe PE  & Jitter\\
    \midrule
    % Ours(down2) & 2.50  & 3.33 & 22.33 & 1.69 & 1.50 & 5.97 &\textbf{2.91} & 9.12 & 2,253.6 & 109.5\\
    % Ours(up4) & \textbf{2.48}  & \textbf{3.30} & 21.67 & 1.56 & \textbf{1.46} & \textbf{5.96} &\textbf{2.91} & 9.01 & 371.1\\
    Parallel Diffusion & 6.73  &14.71 \\
    \rowcolor[RGB]{241, 241, 255} 
    Stratified Diffusion & 6.46 & 10.83 \\
    \bottomrule
  \end{tabular}
  \vspace{-1mm}
  \caption{Evaluation results on the conditional strategy of the diffusion model under setting S1.}
  \label{tab:ablation_cascade}
  \vspace{-4mm}
\end{table}
% \begin{table}[h!]
%   \centering
%   \vspace{-4mm}
%   \resizebox{0.97\linewidth}{!}{
%   \begin{tabular}{@{}c|cccccccc@{}}
%     \toprule
%      Method  & MPJRE & MPJPE & MPJVE & Jitter\\
%     \midrule
%     SAGE(5 Parts) & 2.63 & 3.48 & 20.32 & 7.16\\
%     \rowcolor[RGB]{241, 241, 255} 
%     SAGE(2 Parts) & 2.53 & 3.28 & 20.62 & 6.55\\
%     \bottomrule
%   \end{tabular}
%   }
%   \caption{Comparison between different disentanglement under S1.}
%   \vspace{-4mm}
%   \label{tab:ablation_disentangle}
% \end{table}

\subsection{Dataset and Evaluation Metrics}
We train and evaluate our method on AMASS~\cite{AMASS}, which unifies multiple motion capture datasets~\cite{CMU,BMLrub,HDM05,MPI_Limits,TotalCapture, Eyes_Japan, KIT,BMLmovi, ACCAD, SFU, HumanEval} as SMPL~\cite{SMPL} representations. 
 
 %For evaluating the accuracy of the reconstruction results, we utilize several metrics to assess different aspects. 
 
 We report several metrics for evaluations and comparisons: mean per joint rotation error (MPJRE) and mean per joint position error (MPJPE) for measuring the average relative rotation and position error across all joints respectively, as well as the average position error of the root joints (Root PE), hand joints (Hand PE), upper-body joints (Upper PE), and lower-body joints (Lower PE). 
 
 Besides the above reconstruction accuracy, we also evaluate the spatial and temporal consistency of the generated sequences, as it significantly contributes to the visual quality. Specifically, we calculate the mean per joint velocity error (MPJVE) and Jitter, where MPJVE measures the average velocity error of all body joints, and Jitter quantifies the average jerk (time derivative of acceleration) of all body joints. In both cases, lower values indicate better results. 

 \subsection{Quantitative and Qualitative Results}
For a fair comparison, we follow two settings used in previous works~\cite{FLAG,VAEHMD,Humor,AvatarPoser,AvatarJLM,AGROL} for quantitative and qualitative assessment. Moreover, we propose a new setting in this paper for a more comprehensive evaluation on current methods. 
% To evaluate all methods, we utilize their online inference results rather than offline inference for practical application purposes. 
% We compare our method against baseline methods under these protocols, referring to the reported results from their original papers. 
% For comparison purposes, we train the open source method proposed in ~\cite{AvatarPoser} under protocol 3 and the method in ~\cite{AGROL} under protocol 2, which are not reported by their original paper. We exclude other methods that do not have reported results under specific protocols.

In the first setting, as previous works~\cite{AvatarPoser, AGROL, BoDiffusion, AvatarJLM}, subsets CMU~\cite{CMU}, BMLrub~\cite{BMLrub}, and HDM05~\cite{HDM05} datasets are randomly divided into 90\% for training and 10\% for testing. Besides sparse observations of three joints, we also evaluate the performance of all compared methods by using four joints as input, including the root joint as an additional input, the same as in~\cite{AvatarPoser}. We term this setting as S1 in the following.

\cref{tab:result_p1,tab:result_p1_4input} show that our method outperforms existing methods on most evaluation metrics, confirming its effectiveness. For the MPJVE metric, only AGRoL~\cite{AGROL} surpasses our method when employing an offline strategy. In this scenario, specifically, AGRoL processes the entire sparse observation sequence in one pass and outputs the predicted full-body motions simultaneously. This enables each position in the sequence to utilize the information from both preceding and subsequent time steps, offering an advantage in this particular metric. However, it's important to note that, despite being competitive in metric numbers, offline inference has limited practical applicability in real-world scenarios where online processing capability is most important.

The second setting follows~\cite{VAEHMD, Humor, FLAG, AGROL, AvatarJLM}, where we evaluate the methods on a larger benchmark from AMASS~\cite{AMASS}. The subsets~\cite{SFU,CMU, BMLrub,HDM05, MPI_Limits, TotalCapture, Eyes_Japan, KIT, BMLmovi, KIT, ACCAD, MoSh, SFU} are for training, and Transition~\cite{AMASS} and HumanEva~\cite{HumanEval} subsets are for testing. We term this setting as S2 in the following.

As shown in~\cref{tab:result_p2}, our method achieves comparable performance with previous works on S2. However, we observe that the testing set of S2 is disproportionately small (i.e., only 1\% of the training set). Such a small fraction cannot represent the overall data distribution of the large dataset and may not include sufficiently diverse motions to evaluate the models' scalability, causing unconvincing evaluation results.
%Consequently, the evaluation results are not convincing.
% For instance, in terms of the MPJRE evaluation metric, the gap between state-of-the-art methods (like those in AvatarJLM, AvatarPoser, AGROL) and our approach is around 0.6. Yet, in this benchmark, the performance discrepancy for the same metric is markedly less, at just 0.26.
% We observed that the testing set size is disproportionately small. With only 138 samples for testing compared to 12,925 for training, the test set constitutes a mere 1\% of the training set size. Such a small fraction can undermine the evaluative power of the test set, as it may not adequately represent the overall data distribution or capture the full spectrum of variability inherent in the larger dataset. Consequently, the evaluation results could be skewed, not accurately representing the model's effectiveness on data it has not encountered before.
We introduce a new setting, S3, which adopts the same training and testing splitting ratio used in S1. In this setting, we randomly select 90\% of the samples from the 15 subsets of S2 for training, while the remaining 10\% are for testing. We train and evaluate the compared methods with this new setting.
Table~\ref{tab:result_p3} reveals that under S3, the performance differences between the compared methods are more significant than S1 and S2. 
%This outcome attributes to the application of a larger and more appropriately divided benchmark, which assesses model performance more effectively. 
Since the test set has more diverse motions in S3, this benchmark evaluates the models' scalability in a more objective way. In this context, our method outperforms existing methods in most metrics, especially in the critical metric of Lower PE, highlighting the superiority of our stratified design for lower-body modeling and inference.

\cref{fig:vis} presents a visual comparison between our SAGE Net and baseline methods, all trained under the S1 protocol, which is commonly used by baselines for releasing their trained checkpoints. These visualizations demonstrate the significant improvements that our model offers in reconstructing the lower body. For example, in the first row of samples, baseline methods typically reconstruct the feet too close to the ground, restricting the avatar's leg movements. Our model, however, overcomes this limitation, enabling more flexible leg movements. In the third row, for a subject climbing a ladder, the baseline methods often result in avatars with floating feet, failing to capture the detailed motion of climbing. In contrast, our SAGE Net accurately replicates complex foot movements, resulting in more realistic and precise climbing animations. We also evaluate our model on the real data, and for fair comparison, we directly use the real data release by~\cite{AvatarJLM}. As shown in~\cref{fig:vis_real}, our method also achieves better reconstruction results on the real data. 

% \begin{table*}
%   \centering
%   \begin{tabular}{@{}ccccccccccc@{}}
%     \toprule
%      Method &MPJRE & MPJPE & mpjve & handpe& upperpe & lowerpe & rootpe & jitter\\
%     \midrule
%     Final IK &16.77 & 18.09 &59.24 & - & - & - & - & - \\
%     LoBSTr &10.69 & 9.02 & 44.97 & - & - & - & - & - \\
%     VAR-HMD &4.11 & 6.83 & 37.99 & - & - & - & - & - \\
%     Avatarposer & 3.08 &  4.18 & 27.70 & 2.45 &1.81 & 7.59 & 3.34 & 14.49\\
%     AGRoL(Online) & 3.12 & 4.50 & 107.11 & 1.63& 1.63 & 8.40 & 3.91 \\
%     AvatarJLM & 2.90 & 3.35 & \textbf{20.79} & \textbf{1.24}& \textbf{1.42} & 6.14 & 2.94 & \textbf{8.39}\\
%     % Ours(down2) & 2.50  & 3.33 & 22.33 & 1.69 & 1.50 & 5.97 &\textbf{2.91} & 9.12 & 2,253.6 & 109.5\\
%     % Ours(up4) & \textbf{2.48}  & \textbf{3.30} & 21.67 & 1.56 & \textbf{1.46} & \textbf{5.96} &\textbf{2.91} & 9.01 & 371.1\\
%     Ours & \textbf{2.54} & \textbf{3.26} & 21.99 & 1.36 & \textbf{1.42} &\textbf{5.93} & \textbf{2.91} & 9.29 \\
%     \bottomrule
%   \end{tabular}
%   \caption{Evaluation results under protocol 1.}
%   \label{tab:result_p1}
% \end{table*}

% We argue that the evaluation metrics can not fully evaluate the reconstruction results, better visual results, same/worse evaluation results, explain reason. Tab.~\ref{tab:result_p1}

\subsection{Ablation Study}
We perform ablation study under S1 to justify the design choice of each component in our SAGE Net.

\begin{figure}[t]
    \centering
    \includegraphics[width=0.9\linewidth]{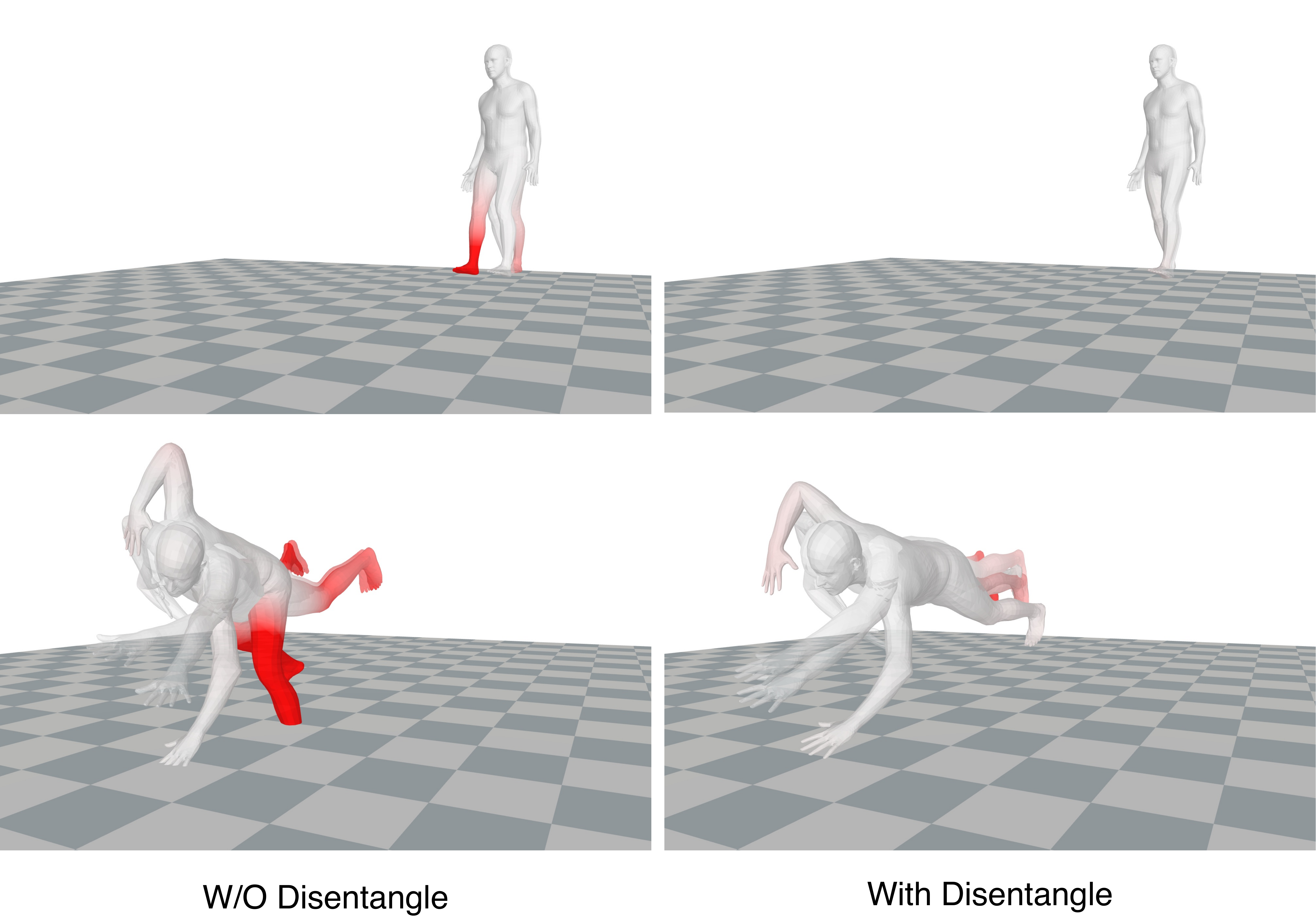}
    \caption{The visualization comparison for disentanglement. The darker the red color, the greater the deviation is between the predicted result and the ground truth.}
    \label{fig:vis_disentangle}
    \vspace{-4mm}
\end{figure}

\myparagraph{Disentangled Codebook:} We establish a baseline using a unified motion representation to evaluate the disentangle strategy. Specifically, we developed a full-body VQ-VAE model that encodes full-body motion into a single, unified discrete codebook. Other components are the same as the original model. Results shown in the first and the last rows in Table~\ref{tab:ablation_disentangle1}, demonstrate that our approach employing disentangled latents significantly outperforms the baseline on all evaluation metrics. This demonstrates that the disentanglement can simplify the learning process by allowing the model to focus on a more limited set of movements and interactions. Additionally, \cref{fig:vis_disentangle} shows the visualization comparison between our model and baseline model, verifying that the disentangle can significantly improve the reconstruction results for the most challenging lower motions.

\myparagraph{Full-Body Decoder and Refiner:} The second and third rows of~\cref{tab:ablation_disentangle1} demonstrate the impact of the full-body decoder and the refiner, respectively. Compared with utilizing the upper and lower decoder from VQ-VAE$_{up}$ and VQ-VAE$_{low}$, the full-body decoder facilitates the integration of features from both the upper and lower body, improving the overall accuracy of full-body motion reconstruction. On the other hand, the refiner acts as a temporal memory, smoothing out the motion sequence to yield better visualization results. 

\myparagraph{Disentanglement Strategy:} To investigate the optimal disentanglement strategy, we explore an extreme disentanglement configuration by following the path from the root (Pelvis) node to each leaf node along the kinematic tree. Specifically, we break down the body into five segments: the paths from the root to the left hand (a), right hand (b), head (c), left foot (d), and right foot (e). As reported in the last two rows of~\cref{tab:ablation_disentangle1}, the natural joint interconnections within the upper (or lower) body were disrupted when further disentangling the human body, resulting in performance drops and complicating the model design.

\myparagraph{Stratified Inference:} \cref{tab:ablation_cascade} highlights the influence of our stratified design on the accuracy of lower body predictions. For comparison, we design a baseline that only uses the sparse observation for lower body latent generation without predicted upper body latent (term as Parallel Diffusion in the table). As we focus solely on the reconstruction quality of the lower body here, we use the decoding results on the generated lower latents from VQ-VAE$_{low}$ to isolate the impact of other modules such as the full-body decoder and refiner. We report Lower PE and Jitter of the lower body for comparison. Results show that our stratified design markedly improves the accuracy of lower body predictions.

\begin{figure}[t]
\vspace{-3mm}
    \centering
    \begin{minipage}[t]{0.3\linewidth}
    \centering
        \includegraphics[width=\linewidth]{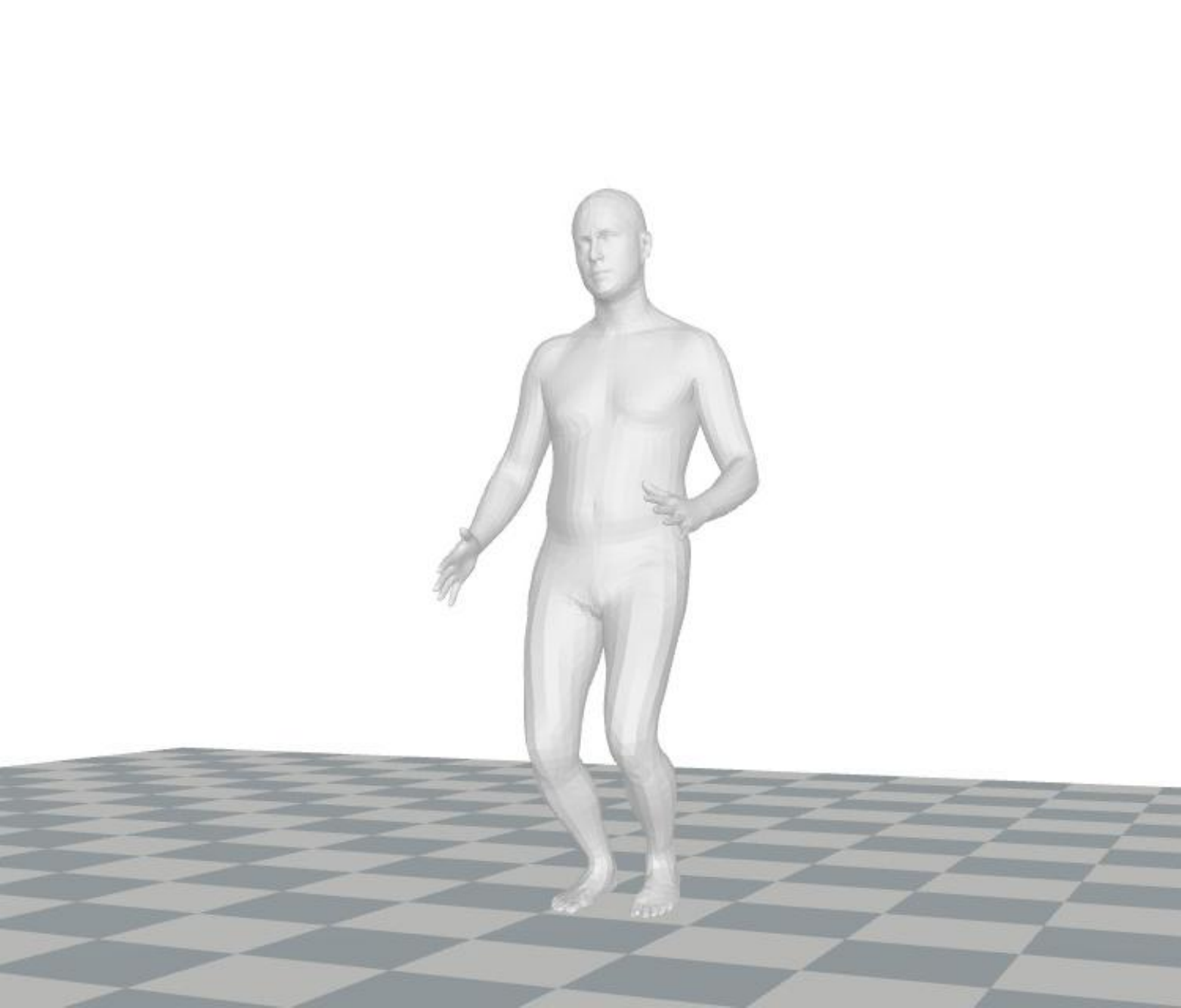}
        \\
        \includegraphics[width=\linewidth]{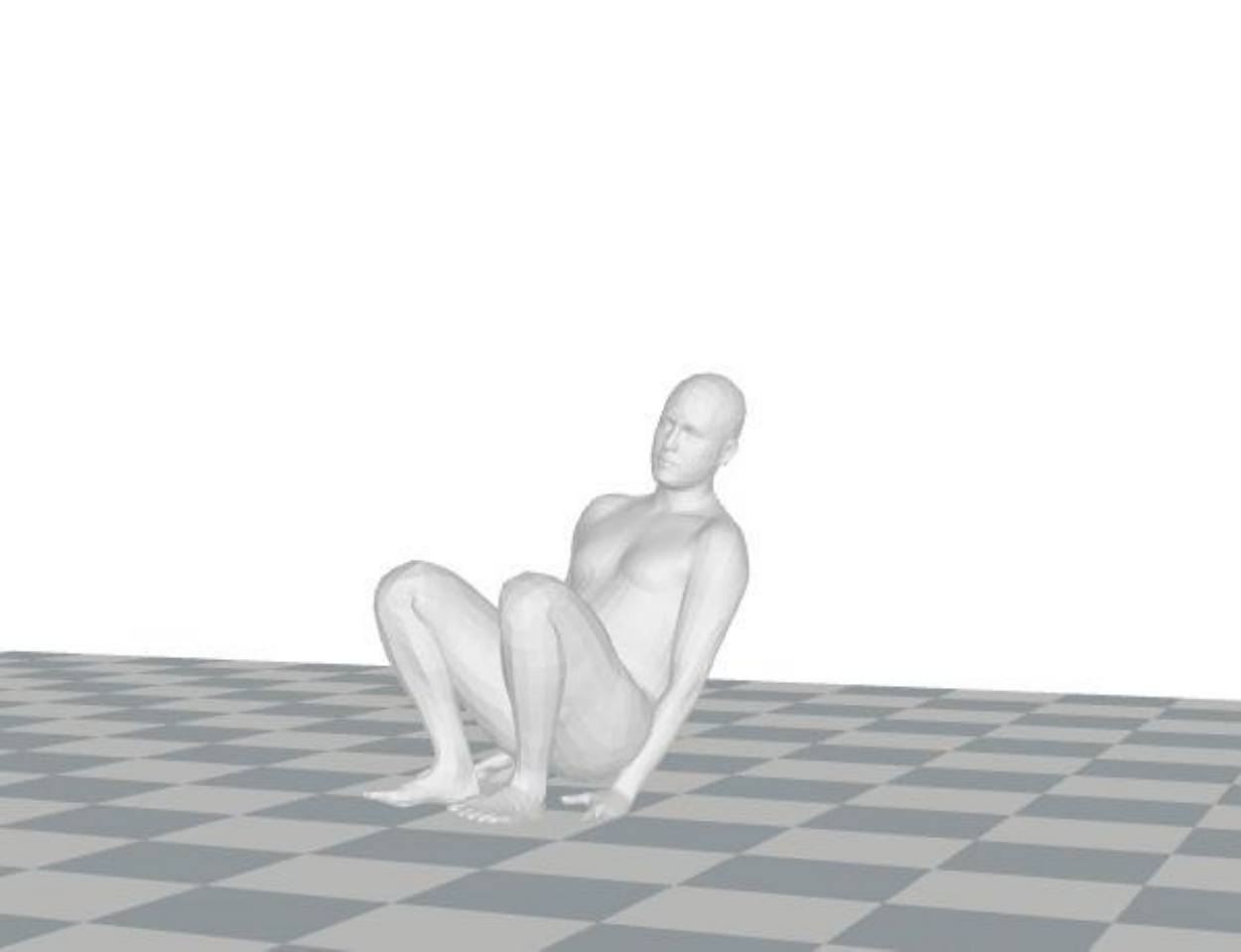} \\
        AvatarJLM~\cite{AvatarJLM}
    \end{minipage}
    \begin{minipage}[t]{0.3\linewidth}
    \centering
        \includegraphics[width=\linewidth]{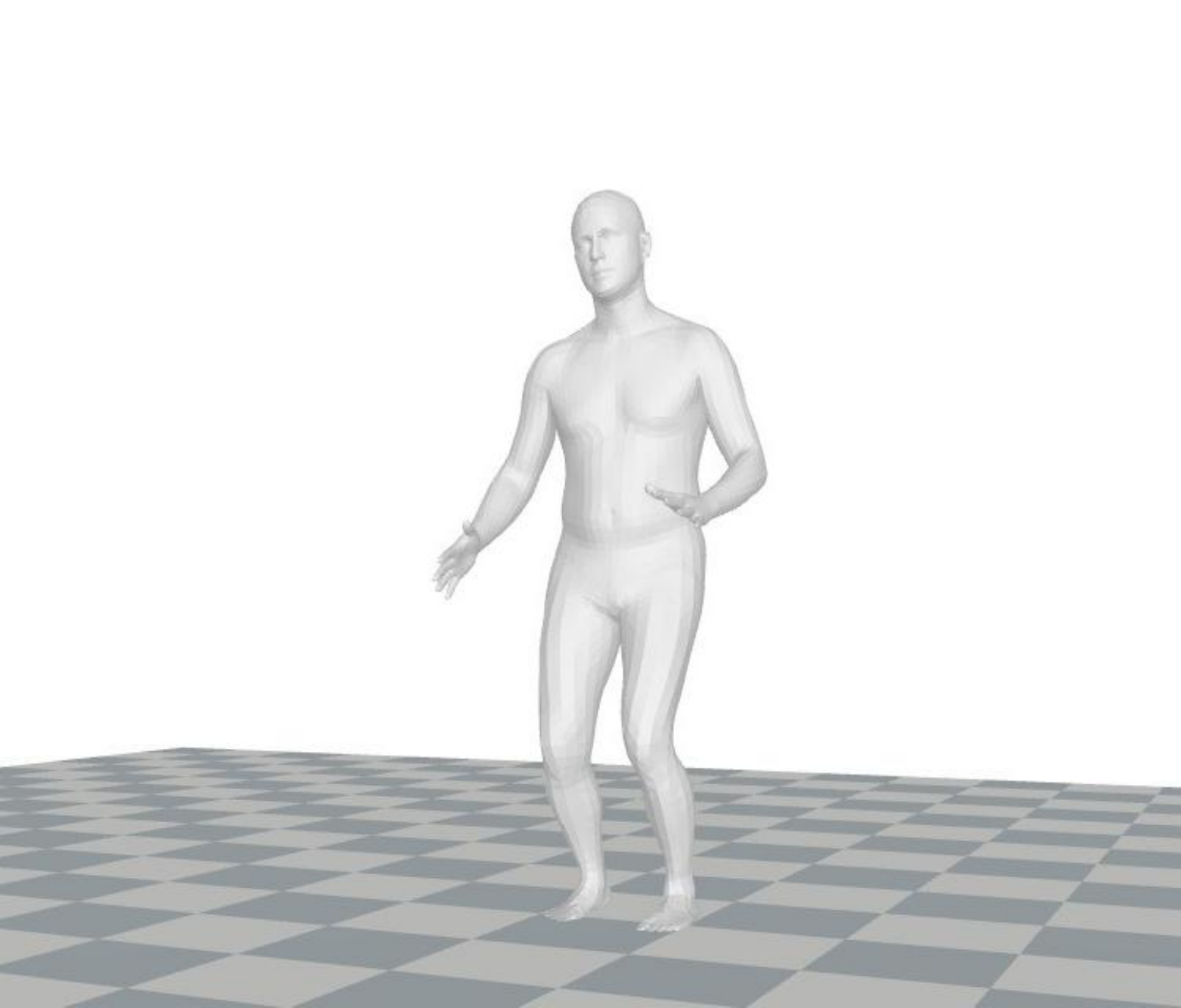} \\
        \includegraphics[width=\linewidth]{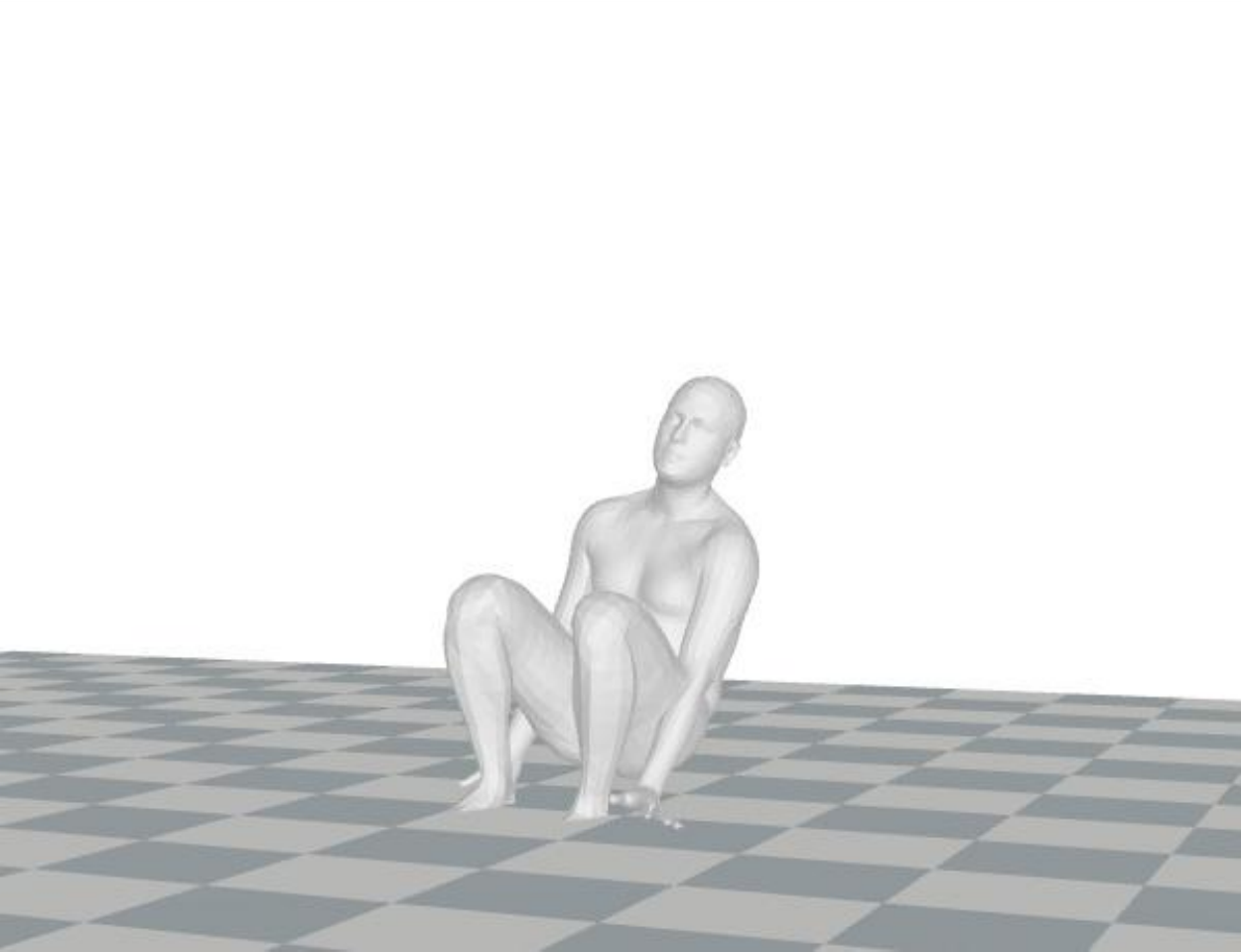} \\
        Ours
    \end{minipage}
    \begin{minipage}[t]{0.3\linewidth}
    \centering
        \includegraphics[width=\linewidth]{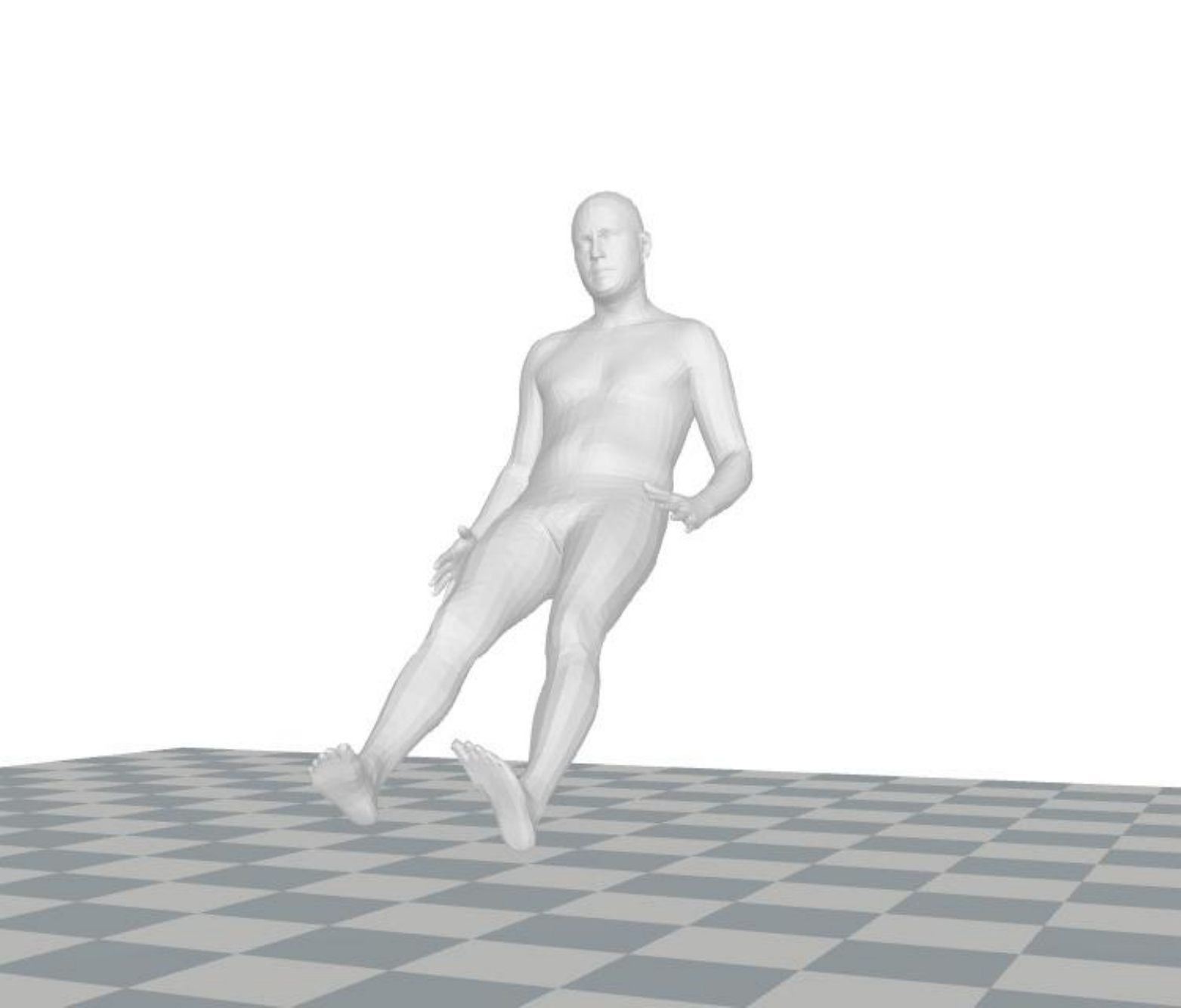} \\
        \includegraphics[width=\linewidth]{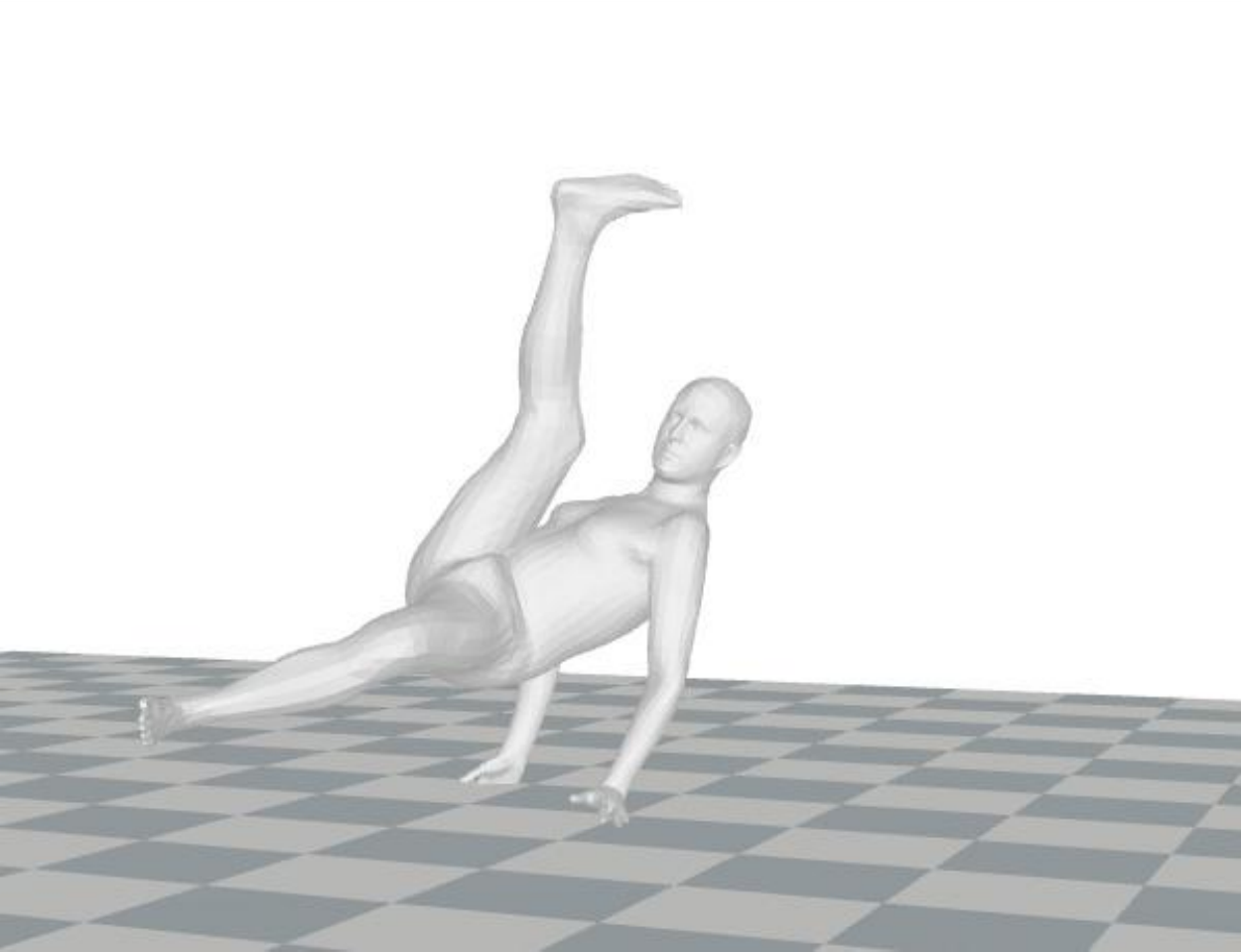} \\
        GT
    \end{minipage}
    \caption{Failure cases. All models are trained under setting S1.}
    \label{fig:failure_cases}
    \vspace{-5mm}
\end{figure}

\myparagraph{Limitation:} In ~\cref{fig:failure_cases}, both the previous state-of-the-art method and our model encounter difficulties in two main situations: (1) External Force-Induced Movements (the top row). (2) Unconventional Poses (the bottom row). The addition of more varied samples to the training dataset can potentially enhance the model’s performance in these areas.

% \textbf{Ablation on codebook size}
% Increasing the size of the codebook won't affect the performance of the model, but affect the performance of the full-body version. Justifying that our disentangled design makes the model more efficient and effective. 

%\section{Conclusion and Limitation}
% \section{Conclusion \& Limitation}
\section{Conclusion}
\label{5_conclusion}

We study the problem of human avatar generation from sparse observations. Our key finding is that the upper and lower body motions should be disentangled with respect to the input signals from the upper-body joints. Based on this, we propose a novel stratified solution where the upper-body motion is reconstructed first, and the lower-body motion is reconstructed next and conditioned on the upper-body motion. Our proposed stratified solution achieves superior performance on public available benchmarks. 
% In this work, we reflect on the drawbacks of current avatar generation methods when predicting complex human motion in a unified space from limited observations. We propose a novel stratified solution where the upper-body motion is reconstructed first, and the lower-body motion is reconstructed next and conditioned on the upper-body motion. Our proposed stratified solution achieves superior performance on public available benchmarks. 

% To potentially enhance the model’s performance in these areas, we are considering the addition of
% more varied samples to the training dataset. 

% {
%     \small
%     \bibliographystyle{ieeenat_fullname}
%     \bibliography{main}
% }
\bibliographystyle{ieeenat_fullname}
\bibliography{main}

\clearpage
\setcounter{page}{1}
\renewcommand{\thetable}{\Alph{table}}
\renewcommand{\thefigure}{\Alph{figure}}
\setcounter{table}{0}
\setcounter{figure}{0}

\maketitlesupplementary

\begin{table*}[h!]
  \centering

  \begin{tabular}{@{}c|c|ccccccc@{}}
    \toprule
     Method & Length & MPJRE & MPJPE & MPJVE & Hand PE& Upper PE & Lower PE & Jitter\\
    \midrule
    AvatarJLM & 10 & 3.19 & 3.76 & 24.67 & 1.31& 1.84 & 7.13 &  11.39\\
    AvatarJLM & 20 & 3.76 & 3.52 & 21.69 & 1.25& 1.73 & 6.65 &  9.17\\
    \rowcolor[RGB]{241, 241, 255} 
    AvatarJLM& 40 & 2.90 & 3.35 & 20.79 & 1.24& 1.72 & 6.20 &  8.39\\
    \midrule
    SAGE (Ours) & 10 & 2.56 & 3.34 & 22.45 & 1.34 & 1.44 & 6.08 &  8.07\\
    \rowcolor[RGB]{241, 241, 255} 
    SAGE (Ours) & 20 & 2.53 & 3.28 & 20.62 & 1.18 & 1.39 & 6.01 &  6.55 \\
    SAGE (Ours) & 40 & 2.51 & 3.20 & 19.36 & 1.39 & 1.43 & 5.75 &  7.28\\
    \bottomrule
  \end{tabular}
  
  \caption{Ablation of the input sequence length. The \textcolor[RGB]{221, 221, 241}{purple} background color denotes the motion length used in the original methods. The computational cost is directly proportional to the length of the input sequence, so we select 20 as our choice for the optimal trade-off between performance and computational cost.}
  \label{tab:ablation_seq}
\end{table*}

\begin{table*}[h!]
  \centering
  \begin{tabular}{@{}c|ccccccccc@{}}
    \toprule
     Method &MPJRE & MPJPE & MPJVE & Hand PE& Upper PE & Lower PE & Root PE & Jitter\\
    \midrule
    SAGE (pred noise) & 3.64 & 4.43 & 25.18 & 3.79 & 2.41 & 7.38 & 3.64 & 9.00\\
    \rowcolor[RGB]{241, 241, 255} 
    SAGE (Ours) & 2.53 & 3.28 & 20.62 & 1.18 & 1.39 & 6.01 & 2.95 & 6.55\\
    \bottomrule
  \end{tabular}
  \caption{Ablation of the diffusion formulation: Predicting original latent $z$ \textit{vs} predicting the residual noise $\epsilon$. Predicting clean latent $z$ achieves superior performance. The \textcolor[RGB]{221, 221, 241}{purple} background color denotes our choice.}
  \label{tab:ablation_noise}
\end{table*}

% \begin{table*}[h!]
%   \centering
%   \begin{tabular}{@{}c|ccccccccc@{}}
%     \toprule
%      DDIM steps & MPJRE & MPJPE & MPJVE & Hand PE& Upper PE & Lower PE & Root PE & Jitter\\
%     \midrule
%     1 & 4.85 & 6.59 & 146.06 & 2.39 & 2.07 & 13.13 & 4.21 & 154.64\\
%     2 & 2.61 & 3.35 & 30.29 & 1.24 & 1.44 & 6.12 & 3.02 & 24.24 \\
%     3 & 2.55 & 3.31 & 22.64 & 1.22 & 1.41 & 6.07 & 2.97 & 11.21 \\
%     4 & 2.54 & 3.29 & 20.93 & 1.19 & 1.40 & 6.02 & 2.95 & 7.49 \\
%     \rowcolor[RGB]{241, 241, 255} 
%     5 & 2.53 & 3.28 & 20.62 & 1.18 & 1.39 & 6.01 & 2.95 & 6.55\\
%     6 & 2.53 & 3.29 & 20.53 & 1.19 & 1.39 & 6.02 & 2.96 & 6.38 \\
%     10 & 2.53 & 3.30 & 20.49 & 1.19 & 1.39 & 6.04 & 2.95 & 6.30\\
%     20 & 2.60 & 3.36 & 20.85 & 1.23 & 1.41 & 6.17 & 2.97 & 6.35\\
%     \bottomrule
%   \end{tabular}
%   \caption{Ablation of the inference sampling steps. The \textcolor[RGB]{221, 221, 241}{purple} background color denotes the optimal inference sampling steps.}
%   \label{tab:ablation_ddim}
% \vspace{-3mm}
% \end{table*}

In this supplementary material, we provide additional ablation on our design choice of the SAGE Net and implementation specifics. 
% In the last, we explore and discuss instances where our SAGE Net does not perform as expected, detailing its failure cases. Additionally, we include a video demonstration that offers a qualitative comparison between our SAGE Net and other baseline methods.

\section*{A. Extra Ablation Studies}
\label{sec:ablationstudy}
\subsection*{A.1 Input sequence length}
Our model adheres to the online inference setting, where it processes sparse tracking signals from the past $N$ frames and predicts the full body motion of the final frame as done in~\cite{AvatarPoser, AvatarJLM}. As indicated in~\cite{AvatarPoser, AGROL, AvatarJLM}, the length of the input sequence is a critical factor affecting the model's performance, involving a balance between efficiency and effectiveness. Therefore, it is essential for our model to effectively tackle shorter sequences, as this not only maintains performance but also significantly reduces computational costs.

% Although our model has demonstrated significant potential in real-time motion prediction, deploying it on AR/VR devices with limited computing power poses a considerable challenge. This challenge is particularly critical as it directly impacts the creation of an immersive user experience.
% One key factor that affects the computational requirements is the length of the input sequence. Therefore, it is crucial for the model to possess the capability to handle shorter sequences effectively. By handling shorter sequences, it can significantly reduce the computational cost. Additionally, it allows the model to leverage future information while maintaining an acceptable level of latency.
We examine AvatarJLM~\cite{AvatarJLM} and our method with different input lengths $N$ under setting S1, as presented in \cref{tab:ablation_seq}. The results demonstrate that our proposed SAGE Net is more robust to variations in the input sequence length compared to the baseline method, AvatarJLM~\cite{AvatarJLM}. Notably, SAGE Net is able to exceed AvatarJLM's performance even when utilizing just a quarter of their sequence length (10 frames for our method compared to 40 frames for AvatarJLM).

\subsection*{A.2 Predicting noise}
\label{subsec:predicting_noise}
Our SAGE Net follows the approach of previous methods~\cite{ramesh22hierarchical, motiondiffusion} by directly predicting the raw data during the diffusion process, specifically the clean latent $z_0$ in our context. In this subsection, we adapt the diffusion process to predict the residual noise $\epsilon$ instead of $z_0$, while maintaining all other components as they are, to validate the effectiveness of this design choice. Results are detailed in \cref{tab:ablation_noise}. We observe that compared with predicting the noise $\epsilon$, this strategy leads to enhanced performance.

% \subsection*{A.3 Inference sampling steps}
% The number of sampling steps required in the inference stage directly impacts computational costs.  To identify the optimal balance between precision and computational efficiency, we conduct ablation studies on the number of sampling steps in DDIM during inference. We aim to determine the minimal number of steps necessary to achieve accurate reconstruction results while keeping computational costs low. As shown in \cref{tab:ablation_ddim}, our method can achieve state-of-the-art performance using only 3-5 denoising steps. 

% More sampling steps at the inference stage require more computational capacity, we conduct ablations on sampling steps on DDIM during inference to investigate the optimal inference steps that can achieve precise reconstruction results while maintain low computational costs. Compared with another diffusion model based method AGROL~\cite{AGROL} that using 50 steps during inference, our method can achieve state-of-the-art performance within 3-5 denoising steps.

% As discussed in \ref{subsec:predicting_noise}, since our diffusion process directly predicts the latent $z$ itself rather than $\epsilon$, it needs only 5 steps to get the best performance.

\section*{B. Implementation Details}
% 要写的东西:
% VQVAE/ first stage/ second stage的结果
% 结构细节等,维度,训练过程
\subsection*{B.1 Disentangled VQ-VAE}
The VQ-VAE$_{up}$ and VQ-VAE$_{low}$ follow the architecture in~\cite{PoseGPT}, unitizing a 4-layer transformer network~\cite{transformer}. Each of these transformer layers includes a 4-head self-attention module and a feedforward layer with 256 hidden units.

For the training of VQ-VAEs, we employ a set of loss terms including a rotation-level reconstruction loss, a forward kinematic loss as proposed in~\cite{AvatarPoser}, and a hand loss as proposed in~\cite{AvatarJLM} with batch size of $512$. Adam optimizer is adapted for training, and we set its Betas parameters to $(0.9, 0.99)$ and the weight decay rate to $1e-4$. The initial learning rate is $1e-4$ and decreases by a factor of 0.2 at the milestone epochs $[25, 35, 50]$.

\subsection*{B.2 Stratified Diffusion}
In our transformer-based model for upper-body and lower-body diffusion, we integrate an additional DiT block as described in~\cite{DiT}. Each model features 12 DiT blocks, each with 8 attention heads, and an input embedding dimension of 512. The full-body decoder is structured with 6 transformer layers.

The diffusion process is trained with 1000 sampling steps, employing the ``squaredcos\_cap\_v2" beta schedule. For this schedule, we set the starting beta value at 0.00085 and the ending beta value at 0.012. The training of the upper-body diffusion model, lower-body diffusion model, and the full-body decoder $D_{full}$, is conducted sequentially. Each component is trained with a batch size of 400, using the Adam optimizer. We set the weight decay at 1e-4 and begin with an initial learning rate of 2e-4. The learning rate undergoes a reduction by a factor of 0.25 at the milestone epochs of 20 and 30.

% During the upper-body diffusion process, the input sparse observation $X$ and noisy latent $z_t^{up}$ are first concatenated along the last dimension, then the concatenated latent features pass through the network to obtain a clean upper-body latent representation $\hat{z^{up}}$. 
% During the lower-body diffusion process, in addition to the input sparse signals $X$ and noisy latent $z_t^{low}$, the clean upper-body latent $\hat{z^{up}}$ generated by the upper-body diffusion process is also incorporated. The concatenated latent features then pass through the lower-body diffusion network to obtain a clean latent representation $\hat{z^{low}}$ for the lower body.
% After the clean upper-body and lower-body latent representations are obtained, instead of directly using pre-trained upper and lower decoders in the VQ-VAE stage, we train a new full-body decoder from scratch together with our stratified motion diffusion to capture the correlations between half-body motion. During the decoder training process, we first freeze the upper-body and lower-body diffusion models to train the decoder, after 15 epochs, we unfreeze the upper-body and lower-body diffusion models to make a thorough end-to-end fine-tune.

\subsection*{B.3 Refiner}
The refiner is a simple two-layer GRU for smoothing the output sequence with minimal 
computational expense. During the training stage, the refiner learns to predict the residual error $\hat{\Theta}_{res}$ between the ground truth motion $\Theta$ and the predicted motion $\hat{\Theta}$ from the full-body decoder. The final rotation prediction $\hat{\Theta}_{final}$ can be obtained by: 
\begin{equation}
    \hat{\Theta}_{final} = \hat{\Theta} + \hat{\Theta}_{res}
\end{equation}
For achieving a balance between smoothness and accuracy in the predicted motion sequences, we adopt various loss terms previously utilized in related research~\cite{AvatarPoser, AvatarJLM}. These include the rotation-level reconstruction loss $L_{rec}$, the velocity loss $L_{vel}$, and the forward kinematic loss $L_{fk}$. 
 
In addition, we design a new loss term jitter loss $L_{jitter}$ to directly control the jitter:
\begin{equation}
\small
        L_{jitter} = \frac{f^3}{N-3}\sum_{i=1}^{i=N-3}||(\hat{v}_{i+2}-\hat{v}_{i+1}) - (\hat{v}_{i+1}-\hat{v}_i)||_2
\end{equation}
where $\hat{v}_i, ~i=1,2...,N-1$, represents the predicted joint velocity of $i^{th}$ frames, and $f$ represents the fps (frames per second).

The complete loss term for training the refiner can be written as: 
$$L = \alpha*L_{rec} + \beta*L_{vel} + \gamma*L_{fk} + \delta*L_{jitter}$$
We set $\alpha, \beta, \gamma, \delta$ to 0.01, 10, 0.05, and 0.01 to force the refiner to focus more on motion smoothness in the training process.

All experiments can be carried out on a single NVIDIA GeForce RTX 3090 GPU card, using the Pytorch framework.

\end{document}